\documentclass[usenames,dvipsnames,sigconf]{acmart}

\usepackage[nolist]{acronym}
\AtBeginDocument{%
 \providecommand\BibTeX{{%
    \normalfont B\kern-0.5em{\scshape i\kern-0.25em b}\kern-0.8em\TeX}}}

\usepackage{soul}
\usepackage{color}
\usepackage{wallpaper}
\usepackage{graphicx}
\usepackage{cleveref}
\usepackage{numprint}
\usepackage{paralist}
\usepackage{xspace}
\usepackage[lofdepth,lotdepth]{subfig}

\usepackage{draftwatermark}

\SetWatermarkText{Preliminary}
\SetWatermarkColor[gray]{0.8}

\definecolor{orangeX}{rgb}{1,.5,0}
\definecolor{blueX}{rgb}{.2, .59, .88}
\definecolor{purpleX}{rgb}{.294118, 0, .509804}
\definecolor{greenX}{rgb}{.421, .578, .241}
\definecolor{bole}{rgb}{0.47, 0.27, 0.23}
\definecolor{mypink3}{cmyk}{0, 0.7808, 0.4429, 0.1412}
\definecolor{mygray}{gray}{0.6}

\newcommand{\hume}{\textsc{Hume-Vidmimic}\,}

\newcommand{\ds}{\textsc{DeepSpectrum\,}}
\newcommand{\wtv}{\textsc{Wav2Vec2.0\,}}
\newcommand{\vit}{\textsc{ViT\,}}
\newcommand{\facenet}{\textsc{FaceNet512\,}}
\newcommand{\opensmile}{\textsc{openSMILE}}

\newcommand{\pyfeat}{\textsc{Py-Feat\,}}

\newcommand{\eg}{e.\,g., }
\newcommand{\ie}{i.\,e., }

\newcommand{\cf}{{cf.\ }}

\begin{document}
\title[MuSe 2023: Baseline Paper]{
The MuSe 2023 Multimodal Sentiment Analysis Challenge: Mimicked Emotions, Cross-Cultural Humour, and Personalisation} %

\author{Lukas Christ}
\affiliation{%
  \institution{EIHW, University of Augsburg}
  \city{Augsburg, Germany}}
  
\author{Shahin Amiriparian}
\affiliation{%
  \institution{EIHW, University of Augsburg}
  \city{Augsburg, Germany}}

\author{Alice Baird}
\affiliation{%
  \institution{Hume AI}
  \city{New York, USA}}

\author{Alexander Kathan}
\affiliation{%
  \institution{EIHW, University of Augsburg}
  \city{Augsburg, Germany}}
  
  \author{Niklas Müller}
\affiliation{%
  \institution{University of Passau}
  \city{Passau, Germany}}
  
  \author{Steffen Klug}
\affiliation{%
  \institution{University of Passau}
  \city{Passau, Germany}}

\author{Chris Gagne}
\affiliation{%
  \institution{Hume AI}
  \city{New York, USA}}

\author{Panagiotis Tzirakis}
\affiliation{%
  \institution{Hume AI}
  \city{New York, USA}}
  
\author{Eva-Maria Meßner}
\affiliation{%
  \institution{University of Ulm}
  \city{Ulm, Germany}}
  
\author{Andreas König}
\affiliation{%
  \institution{University of Passau}
  \city{Passau, Germany}}
  
\author{Alan Cowen}
\affiliation{%
  \institution{Hume AI}
  \city{New York, USA}}
  
\author{Erik Cambria}
\affiliation{%
  \institution{Nanyang Technological University}
  \city{Singapore, Singapore}}

\author{Bj\"orn W. Schuller}
\affiliation{%
  \institution{GLAM, Imperial College London}
  \city{London, United Kingdom}}
\renewcommand{\shortauthors}{Lukas Christ et al.}

\settopmatter{printacmref=true}
\copyrightyear{2023}
\acmYear{2023}
\setcopyright{acmlicensed}

\begin{abstract}

The \ac{MuSe} 2023 is a set of shared tasks addressing three different contemporary multimodal affect and sentiment analysis problems: In the \ac{MuSe-Mimic}, participants predict three continuous emotion targets. This sub-challenge utilises the \hume dataset comprising of user-generated videos. For the \ac{MuSe-Humor}, an extension of the \ac{Passau-SFCH} dataset is provided. Participants predict the presence of spontaneous humour in a cross-cultural setting. The \ac{MuSe-Personalisation} is based on the \ac{Ulm-TSST} dataset, featuring recordings of subjects in a stressed situation. Here, arousal and valence signals are to be predicted, whereas parts of the test labels are made available in order to facilitate personalisation.
\ac{MuSe} 2023 seeks to bring together a broad audience from different research communities such as audio-visual emotion recognition, natural language processing, signal processing, and health informatics.
 In this baseline paper, we introduce the datasets, sub-challenges,  and provided feature sets. As a competitive baseline system, a \ac{GRU}-\ac{RNN} is employed. On the respective sub-challenges' test datasets, it achieves a mean (across three continuous intensity targets)
 Pearson's Correlation Coefficient of  $.4727$ for \ac{MuSe-Mimic}, an \ac{AUC} value of $.8310$ for \ac{MuSe-Humor} and \ac{CCC} values of $.7482$ for arousal and $.7827$ for valence in the \ac{MuSe-Personalisation} sub-challenge.

\end{abstract}

\begin{CCSXML}
<ccs2012>
<concept>
<concept_id>10010147.10010257.10010293.10010294</concept_id>
<concept_desc>Computing methodologies~Neural networks</concept_desc>
<concept_significance>500</concept_significance>
</concept>
<concept>
<concept_id>10010147.10010178</concept_id>
<concept_desc>Computing methodologies~Artificial intelligence</concept_desc>
<concept_significance>500</concept_significance>
</concept>
<concept>
<concept_id>10010147.10010178.10010224</concept_id>
<concept_desc>Computing methodologies~Computer vision</concept_desc>
<concept_significance>300</concept_significance>
</concept>
<concept>
<concept_id>10010147.10010178.10010179</concept_id>
<concept_desc>Computing methodologies~Natural language processing</concept_desc>
<concept_significance>300</concept_significance>
</concept>
</ccs2012>
\end{CCSXML}

\ccsdesc[500]{Computing methodologies~Neural networks}
\ccsdesc[500]{Computing methodologies~Artificial intelligence}
\ccsdesc[300]{Computing methodologies~Computer vision}
\ccsdesc[300]{Computing methodologies~Natural language processing}

\keywords{Multimodal Sentiment Analysis; Affective Computing; Humour Detection; Emotion Recognition; Multimodal Fusion; Challenge; Benchmark}

\maketitle
\section{Introduction}
In its 4th edition, the \textbf{Mu}ltimodal \textbf{Se}ntiment Analysis (MuSe) Challenge proposes three different tasks, namely  categorical emotion prediction,
cross-cultural humour detection,  and personalised dimensional emotion regression. 
For the Emotional Mimicry Sub-Challenge (\textbf{\ac{MuSe-Mimic}}), emotional mimics are explored by introducing a first-of-its-kind, large-scale (557 subjects, 30+ hours), multimodal (audio, video, and text) dataset. The data were gathered in the wild, with subjects recording their own facial and vocal mimics to a wide range of `seed' videos via their webcam. Subjects selected the emotions they perceived in each video out of 63 provided categories plus ``no person can be seen or heard in this video'' and rated each selected emotion on a 0--100 intensity scale. In this sub-challenge, participants will apply a multi-output regression to predict the intensities of three self-reported emotions from the subjects' multimodal recorded responses related to decision-making emotional categories: \begin{inparaitem}[]\item Approval, \item Disappointment,  and \item Uncertainty  \end{inparaitem}. 

For the 
\acl{MuSe-Humor} (\textbf{\ac{MuSe-Humor}}), participants will train their models to predict humour in German football press conference recordings. Different from the training data, the unseen test set consists of videos of English football press conferences, thus providing a cross-cultural evaluation setting.  The \acl{Passau-SFCH} (\textbf{\ac{Passau-SFCH}}) dataset~\cite{christ2022multimodal} as featured in MuSe 2022~\cite{xu2022hybrid, chen2022integrating, li2022hybrid} is provided as the training set. For the test set, we extend \ac{Passau-SFCH} with recordings of press conferences given by 7 different coaches from the English Premier League between September 2016 and September 2020. Both the training and the test set only contain segments in which the respective coach is speaking, accounting for more than 17 hours of data in total. While all provided videos are originally labelled according to the \ac{HSQ} proposed by~\citet{martin2003individual}, the prediction target in \ac{MuSe-Humor} is binary, \ie presence or absence of humour.

In the further featured  \acl{MuSe-Personalisation} (\textbf{\ac{MuSe-Personalisation}}), participants predict continuous estimations of valence and arousal using personalised approaches. 
Different from the usual speaker-independent challenge setup employed in recent years, participants will also be provided with labelled data of each subject from the test partition. 
Thus, \ac{MuSe-Personalisation} encourages the exploration of the adaptation of multimodal emotion recognition systems to individuals, taking their specific features into account.
\ac{MuSe-Personalisation} utilises the \acl{Ulm-TSST} dataset (\ac{Ulm-TSST}) introduced in MuSe 2021~\cite{stappen2021muse, stappen2021summary}. \ac{Ulm-TSST} consists of recordings of 69 individuals undergoing the \ac{TSST}, a scenario designed to induce stress. Besides audio-visual recordings and their textual transcripts, \ac{Ulm-TSST} includes the subjects' EDA, Electrocardiogram (ECG), Respiration (RESP), and heart rate (BPM) signals, each of them sampled at a rate of 1\,kHz.

\begin{table}[t!]
\footnotesize
  \caption{ Statistics on the datasets of each sub-challenge. Given are the number  of unique subjects(\#), and the video durations in the format  hh\,:mm\,:ss. Note that for \ac{MuSe-Personalisation}, partial recordings of the test subjects are also included in the train and development partition, as denoted by the round brackets.
\label{tab:partitioning}
 }
 \resizebox{\linewidth}{!}{
  \begin{tabular}{lrcrcrc}
    \toprule
     & \multicolumn{2}{c}{\textbf{\ac{MuSe-Humor}}} & \multicolumn{2}{c}{\textbf{\ac{MuSe-Mimic}}} & \multicolumn{2}{c}{\textbf{\ac{MuSe-Personalisation}}}\\
     \cmidrule(lr){2-3} \cmidrule(lr){4-5} \cmidrule(lr ){6-7}
    Partition & \# & Duration & \# &  Duration & \# & Duration \\
    \midrule
    Train   & 7 & 7\,:44\,:49 & 328 &19\,:24\,:23& 41 (+14) & 3\,:39\,:56 \\
    Development  &  3 & 3\,:06\,:48 & 107 & 6\,:22\,:22& 14 (+14) & 1\,:20\,:10  \\
    Test    &  6 & 6\,:35\,:16 & 122 & 6\,:33\,:10 & 14 & 0\,:47\,:21 \\
    \midrule
    $\sum$    & 16 & 17\,:26\,:53 & 557 & 32\,:19\,:56 & 69 & 5\,:47\,:27 \\
  \bottomrule
\end{tabular}
}
\end{table}

With their variety of data and prediction targets, the three sub-challenges target a broad audience, including but not limited to researchers interested in affective computing, multimodal representation learning, natural language processing, machine learning, and signal processing. \ac{MuSe} is intended to serve as a common forum to compare different approaches to the proposed tasks, thereby leading to novel insights into the aptitude of different methods, modalities, and features.

In~\Cref{sec:challenges}, the mentioned sub-challenges, their corresponding datasets, and the challenge protocol are outlined in more detail. Next,~\Cref{sec:features} reports our pre-processing and feature extraction pipeline, as well as the experimental setup used to compute baseline results for each sub-challenge. The results are then presented in~\Cref{sec:results}, before~\Cref{sec:conclusion} concludes the paper.

\section{The Three Sub-Challenges}\label{sec:challenges}
Following, we provide more details on each sub-challenge and dataset as well as the challenge protocol.

\subsection{The MuSe-Mimic Sub-Challenge\label{sec:mimic}}

The acquisition of data pertaining to human expressive behaviour remains a challenging task for researchers in the field of affective computing. To address this, researchers have employed various strategies, including the technique of mimicking human expressions in real-world settings~\cite{brooks2023deep}. The utilisation of this approach allows for specific emotions to be targeted with greater efficiency. Emotional mimicry has been observed in childhood development and typically occurs spontaneously during social interactions~\cite{hess2013emotional}. Furthermore, it has been demonstrated that humans possess a highly precise ability to mimic emotions, even down to specific features~\cite{hess2014emotional}.

To address the challenge of acquiring data related to human expressive behaviour, a multimodal dataset of mimics, called the \hume dataset, has been made available for use in the \ac{MuSe-Mimic} sub-challenge. This dataset includes over 30 hours of video data, with a mean duration of 6.4 seconds, collected from 557 subjects located in the United States (cf.~\Cref{tab:partitioning}, for partitioning details). The subjects in the dataset range in age from 19 to 59 years and were recruited via Prolific, with reimbursement for their time.

In the dataset, each subject was instructed to mimic a seed video featuring someone expressing emotion, and then rate the intensity of the seed video, choosing from a selection of emotional classes. The emotional expressions of Approval, Disappointment, and Uncertainty are targeted for this sub-challenge. The labels of these three continuous intensity targets were determined through an agglomerative clustering approach applied to a filtered correlation matrix of all 63 available labels\footnote{Admiration, Adoration, Aesthetic Appreciation, Amusement, Anger, Annoyance, Anxiety, Approval, Awe, Awkwardness, Boredom, Calmness, Concentration, Confidence, Confusion, Contemplation, Contempt, Contentment, Craving, Determination, Disappointment, Disapproval, Discomfort, Disgust, Distress, Doubt, Ecstasy, Embarrassment, Empathic Pain, Enthusiasm, Entrancement, Envy, Excitement, Fear, Frustration, Gratitude, Guilt, Hesitancy, Horror, Interest, Joy, Love, Nostalgia, Pain, Panic, Pride, Realisation, Relief, Resentment, Romance, Sadness, Sarcasm, Satisfaction, Serenity, Sexual Desire, Shame, Shock, Surprise (negative), Surprise (positive), Sympathy, Tiredness, Triumph, and Uncertainty.} (plus no person seen or heard in the video) with $\rho \geq 0.3$. 
For each sample, 
after removing from the 63 emotions the ones with a lower correlation and normalisation of the intensity per emotion, the target label was determined by taking the mean intensity rating for a seed-mimic pair, across the corresponding cluster of labels. The clustering is as follows: 
Approval unites Approval, Admiration, Adoration, Enthusiasm, Excitement, Joy, and Love; Uncertainty unites Uncertainty, Doubt, Romance, Sexual Desire, Shock, and Surprise (negative); and finally, Disappointment unites Disappointment, Anger, Annoyance, Contempt, Disapproval, Disgust, and Frustration.

The data preparation process involved a speaker-independent partitioning of the dataset into training, development, and test sets. 
An automatic transcription is provided for each sample, and the faces of individuals within the videos were detected at a frequency of 2 frames per second.

For the \ac{MuSe-Mimic} sub-challenge, the aim is to perform a multi-output regression from features extracted from the multimodal (audio, video, and text) data for the intensity of the 3 emotional intensity targets. Pearson's correlation coefficient  ($\rho$) is used as the evaluation metric.

.

\subsection{The MuSe-Cross-Cultural Humour Sub-Challenge\label{sec:humor}} 
Humour -- defined as an expression that establishes surprising or incongruent relationships or meaning with the intent to amuse~\cite{gkorezis2014leader}--constitutes one of the most complex phenomena in human social interaction with manifold potential positive or negative effects~\cite{cann2014assessing}. Thus, humour has been a focal research interest in affective computing and human-computer interaction, such as natural language interfaces ~\cite{binsted1995using}. As humour can be expressed both verbally and non-verbally, multimodal approaches are especially suited for detecting humour. Several datasets for multimodal humour recognition have been proposed~\cite{hasan2019ur, mittal2021so, wu2021mumor}, but typically rely on staged scenarios such as TED talks or TV series. Also, some studies equate audience laughter with (successful) humour which is a common, yet crucial flaw. In contrast, to the best of our knowledge, \ac{Passau-SFCH} is the only database for predicting spontaneous and non-scripted displays of humour with a nuanced humour measurement. 

Since humour is embedded in linguistic and contextual factors, a cross-cultural study can help to understand commonalities and differences in humour usage. 
Recently, studies are diving into the multimodal intricacies of humour in different countries, such as differences in displayed smiling behaviours of American and French persons ~\cite{priego2018smiling} or amongst others, gesture and prosody for humour construction in German-Brazilian interactions ~\cite{ladilova2022humor}. %
However, to the best of our knowledge, automated multimodal cross-cultural humour detection, which sheds light on the transferability of humour, has not been done, yet. \ac{MuSe-Humor} is designed to elicit first insights for this topic.

Participants will train their models utilising \ac{Passau-SFCH}'s press conference recordings of 10 different German Bundesliga football coaches. The English test set comprises press conference recordings of 6 football coaches from the English Premier League. However, only one of them is a native English speaker, while the other five coaches come from $5$ different countries (Argentine, France, Germany, Portugal, and Spain). Every coach in both the training/development and the test set is male. The training subjects are aged between 30 and 53 years, and the test subjects' age span ranges from 47 to 57 years. We split the German coaches into a training and development partition for our baseline experiments, where the development partition is identical to that of 2022's \ac{MuSe-Humor} challenge~\cite{christ2022muse, amiriparian2022muse}. Detailed statistics on the dataset can be found in~\Cref{tab:partitioning}.

 We segment all videos such that the data only includes the parts in which the coach is actually speaking.
 Both training and test data include the segmented audio-visual recordings as well as manual transcripts with timestamps. 
All videos are initially annotated at a 2\,Hz rate for the sentiment and direction (self-directed vs others-directed) of humorous utterances, following the \ac{HSQ}~\cite{martin2003individual}. We obtain binary labels denoting the presence or absence of humour as described in~\cite{christ2022muse}, leading to 2\,s video frames, each of which is either considered humorous or not.
In total, 4\,38\,\% of the training data, 2\,81\% of the development data, and 6\,17\% of the test data are labelled as humorous in the gold standard.

Analogously to 2022's \ac{MuSe-Humor} sub-challenge, we employ the \ac{AUC} metric as the evaluation criterion.

\subsection{The MuSe-Personalisation Sub-challenge}\label{ssec:personalisation}

When working with real-world data, often significant differences can be observed between individuals (\eg high variance in pitch within a person's speech or variations in personality characteristics such as genders, age ranges, or cultural background)~\cite{weisberg2011gender,akyunus2021age,li2023survey}. However, most approaches today tend to neglect these individual variations, resulting in models trained on a broad population which do not always generalise well to subjects not present in the training set~\cite{doddington1998sheep}. To solve this problem, personalisation methods are needed that incorporate the characteristics of one individual, leading to personalised models which are capable of providing more accurate predictions~\cite{li2023survey}. \ac{MuSe-Personalisation} is aimed to serve as a benchmark for personalisation in multimodal affect analysis.

Participants of this sub-challenge are supplied with the multimodal \ac{Ulm-TSST} dataset. It consists of recordings of subjects participating in the \ac{TSST}~\cite{kirschbaum1993trier}, which defines a stress-inducing, job interview-like scenario. Participants are asked to deliver a five-minute free speech presentation on why they are suited for a hypothetical job. \ac{Ulm-TSST} comprises recordings of these speeches by 69 subjects (age range 18-39), 49 of which are female, accounting for circa 6 hours of video data.

The data is partitioned into a training, development, and test set in a speaker-independent manner. The training dataset comprises 41, both development and test sets 14 videos. Note that the split is identical to the splits of Ulm-TSST employed for 2021's and 2022's \ac{MuSe-Stress} challenges~\cite{stappen2021muse, christ2022muse}. In order to facilitate personalisation on the test subjects, we provide labelled parts of their videos as follows: we take the first 60 seconds of each test video and consider this our subject-specific training data. Moreover, the next 60 seconds are employed as subject-specific development set. The remaining parts of each test subject's video serve as the sub-challenge's test data, \ie their annotations are kept confidential. \Cref{tab:partitioning} lists key statistics of this partitioning of \ac{Ulm-TSST}.

\ac{Ulm-TSST} has been labelled with 2\,Hz arousal and valence signals by three annotators. The gold standards are identical to those in 2022's \ac{MuSe-Stress} challenge~\cite{christ2022muse} which attracted considerable interest, \eg~\cite{yadav2022comparing, park2022towards, liu2022improving, he2022multimodal}. For the valence gold standard, we fuse these three annotations employing the \ac{RAAW} method~\cite{stappen2021toolbox}.
\ac{RAAW} combines temporal alignment utilising \ac{CTW}~\cite{zhou2015generalized} with annotation fusion via \ac{EWE}~\cite{grimm2005evaluation}.
Regarding arousal, for each video, the signals of the two annotators with the highest agreement are merged with the video subject's downsampled and smoothed EDA signal, as EDA has been shown to be an objective indicator of arousal~\cite{caruelle2019use}, in contrast to inevitably subjective annotations. Details on the gold standard creation can be found in~\cite{christ2022muse}, for extensive experiments on merging biosignals and annotation signals for arousal see~\cite{baird2021physiologically}.

We provide participants with the audiovisual recordings, manual transcripts and the ECG, RESP, and BPM signals.

\subsection{Challenge Protocol}
In order to enter the challenge, participants need to hold an academic affiliation and complete the EULA available on the \ac{MuSe} 2023 homepage\footnote{\href{https://www.muse-challenge.org}{https://www.muse-challenge.org}}. The organisers do not take part in any sub-challenge as competitors. 
During the competition, participants submit their predictions for test labels on the CodaLab platform\footnote{ \href{https://codalab.lisn.upsaclay.fr/}{https://codalab.lisn.upsaclay.fr/}. The link(s) to the respective challenges will be emailed to the participants.}. Up to 5 predictions are possible in each sub-challenge. All participating teams are encouraged to submit a paper describing their experiments and results. In order to officially win a sub-challenge, a paper, which must be accepted, is mandatory. Papers will undergo a double-blind peer-reviewing process.

\section{Baseline Approaches}\label{sec:features}
We supply an extensive set of pre-extracted features in order to support participants in time-efficient model development. For each sub-challenge, seven feature sets (3 audio-based, 3 video-based, 1 text-based), as outlined in the following, are provided\footnote{Note: Participants may employ other external resources (\eg features, datasets, pretrained foundation models). The usage of additional resources, tools etc. is expected to be clearly stated in the corresponding paper.}.

\subsection{Pre-processing}
As described in~\Cref{sec:challenges}, we partition every sub-challenge's dataset into training, development, and test sets. In doing so, we take length, speaker independence, and label distributions into account (\cf~\Cref{tab:partitioning}). 
For \ac{MuSe-Mimic} -- as can be seen in \Cref{tab:partitioning} --, a 60-20-20\% split strategy is applied. There is no additional segmentation applied. Each sample contains a single mimic of a seed video, and labels were normalised per sample to range from [0\,:1]. 

The \ac{Passau-SFCH} press conference recordings are segmented into clips in which the respective coach is speaking. We manually remove a few of these clips where the coach is using any language other than English. Moreover, we discard clips whose audio quality is considerably impaired. 

The recordings in the \ac{Ulm-TSST} dataset are cut to only show the job interview presentation defined by the \ac{TSST} protocol. To preserve subjects' privacy, we also delete parts of videos in which they disclose their names. Besides the segmentation of the test subjects' videos described in \Cref{ssec:personalisation}, no further segmentation is conducted for \ac{MuSe-Personalisation}.

\subsection{Audio}
Before extracting audio features, we normalise all audio files to -3 decibels and convert them to mono, at 16\,kHz, 16\,bit. We then utilise the \opensmile{}~\cite{eyben2010opensmile} toolkit to compute handcrafted features. Moreover, we compute high-dimensional audio representations using both \ds~\cite{Amiriparian17-SSC} and a variant of \wtv~\cite{baevski2020wav2vec}. 

Both systems have proved valuable in audio-based \ac{SER} tasks~\cite{Amiriparian22-DAP,Gerczuk22-EAT,Schuller21-TI2}.

\subsubsection{\acs{eGeMAPS}}
\label{ssec:egemaps}
Using the \opensmile{} toolkit~\cite{eyben2010opensmile}\footnote{\href{https://github.com/audeering/opensmile}{https://github.com/audeering/opensmile}}, we extract  88 \ac{eGeMAPS} features~\cite{eyben2015geneva} which have shown their suitability for sentiment analysis and \ac{SER} tasks~\cite{baird2019can,vlasenko2021fusion}. For each sub-challenge, we apply the standard configuration and extract features with a window size of two seconds, and a hop size of 500\,ms.

\subsubsection{\ds}

We apply \ds~\cite{Amiriparian17-SSC}\footnote{\href{https://github.com/DeepSpectrum/DeepSpectrum}{https://github.com/DeepSpectrum/DeepSpectrum}} to obtain deep \ac{CNN}-based representations from audio data. In particular,  we first create a Mel-spectrogram ($128$ Mels, \emph{viridis} colour mapping) from each audio file with a window size of one second and a hop size of $500$\,ms. We then forward the spectrogram representation into \textsc{DenseNet121} and take the output of the last pooling layer as a $1\,024$-dimensional feature vector. The efficacy of \ds{} has been shown for speech and audio recognition tasks~\cite{Ottl20-GSE, Amiriparian17-SAU, Amiriparian20-TCP}.

\subsubsection{\wtv}

Self-supervised pretrained Transformer models have attracted considerable interest in computer audition recently~\cite{ssl-survey}. A popular example of such a foundation model is \wtv~\cite{baevski2020wav2vec}, which has frequently been employed for \ac{SER}~\cite{pepino21_interspeech, morais2022speech}. As all sub-challenges are affect-related, we utilise a large version of \wtv fine-tuned on the MSP-Podcast~\cite{Lotfian_2019_3} dataset for continuous emotion recognition~\cite{wagner2022dawn}\footnote{\href{https://huggingface.co/audeering/wav2vec2-large-robust-12-ft-emotion-msp-dim}{https://huggingface.co/audeering/wav2vec2-large-robust-12-ft-emotion-msp-dim}}. We extract features for an audio signal by averaging over its representations in the final layer of this model, resulting in 1024-dimensional embeddings.
Regarding \ac{MuSe-Humor} and \ac{MuSe-Personalisation}, we obtain 2\,Hz features by sliding a 3\,s window over each audio file, with a step size of 500\,ms. Due to the comparably short segments in \ac{MuSe-Mimic}, a 2\,s window size is applied there.

\subsection{Video}
We compute features of the visual modality based on the subjects' faces. Therefore, we first automatically extract faces via \ac{MTCNN}. Subsequently, \acp{FAU}, \facenet and \vit representations are obtained for them.

\subsubsection{\acs{MTCNN}}
 
We employ the \ac{MTCNN}~\cite{zhang2016mtcnn} face detection model\footnote{\url{https://github.com/ipazc/mtcnn}}, to extract pictures of the subjects' faces. 

In the videos of \ac{Passau-SFCH}, typically several persons can be seen, whereas only the coach is relevant. In the first step, we attempt to filter out the respective coach's face per video automatically via clustering of face embeddings. Afterwards, we manually correct the thus obtained face sets and only keep the one corresponding to the coach. For \ac{Ulm-TSST} and \hume, no such postprocessing is necessary. In both datasets' recordings, typically only one person is displayed per video.
Subsequently, the extracted face images serve as the basis for feature extraction via \pyfeat, \facenet, and \vit.

\subsubsection{FAU}The concept of \acp{FAU}~\cite{ekman1978facial} provides an interpretable way of encoding facial expressions by reference to the activation of certain facial muscles. 
As facial expressions contain important cues to a person's affective state, \acp{FAU} have received considerable attention from the AC community~\cite{zhi2020comprehensive}. 
We compute automatic estimations of the activation of 20 different \acp{FAU} via the respective SVC model included in the \pyfeat\footnote{\url{https://py-feat.org}} library.

\subsubsection{\facenet}
In order to compute high-dimensional face representations, we make use of the \facenet model~\cite{schroff2015facenet} as implemented in the deepface library~\cite{serengil2020lightface}. \facenet is trained on the task of face recognition and  yields a 
512-dimensional embedding for every face.

\subsubsection{Vision Transformer (\vit)}

As an alternative vision-based strategy, we employ the DINO-trained \vit, which has been pre-trained on the ImageNet-1K dataset in a self-supervised manner using the self-distillation with no labels (DINO) method\cite{caron2021emerging}. This model has demonstrated its efficacy for various image-based tasks, including emotion recognition from facial expressions~\cite{chaudhari2022vitfer, vaiani2022viper}. The model processes the extracted facial images and outputs a 384-dimensional embedding for each image. We do not conduct any further pretraining or fine-tuning.

\subsection{Text: Transformers}

Text-based features are obtained via pre-trained Transformer models in the fashion of \textsc{BERT}~\cite{devlin-etal-2019-bert}. We compute sentence representations by taking the model's final layer's encoding for the \verb|CLS| token, a special token referring to the whole input sentence. For \ac{MuSe-Mimic}, the English transcripts are encoded via a small pretrained \textsc{ELECTRA}~\cite{clark2020electra} model, yielding $256$-dimensional embeddings.
As \ac{MuSe-Humor} features a German training and development, but an English test set, we opt for the multilingual version of BERT~\cite{devlin-etal-2019-bert}\footnote{\href{https://huggingface.co/bert-base-multilingual-cased}{https://huggingface.co/bert-base-multilingual-cased}} here. This model has been pretrained on Wikipedia entries in 104 languages including German and English. It has been shown to generalise effectively from one language to another~\cite{pires-etal-2019-multilingual}. Regarding the German-only transcripts for \ac{MuSe-Personalisation}, we employ a German BERT model\footnote{\href{https://huggingface.co/dbmdz/bert-base-german-cased}{https://huggingface.co/dbmdz/bert-base-german-cased}}. Both mentioned \textsc{BERT} variants account for $768$-dimensional representations. Note that we do not fine-tune any of the models.

\subsection{Alignment}
Each dataset includes audio, video, and transcripts. In order to ease the development of multimodal models utilising the provided features, we align the different features along with each other and the respective task's labelling scheme.

For all tasks, both the audio and the face-based features are extracted with a rate of 2\,Hz by employing sliding windows of step size 500\,ms for audio and sampling faces at a 2\,Hz rate for video. 
In order to obtain sentence-wise timestamps for \ac{MuSe-Humor} and \ac{MuSe-Personalisation} from the manual transcripts, we employ a pipeline consisting of three steps: first, we utilise the Montreal Forced Aligner (MFA)~\cite{mcauliffe2017montreal} toolkit to generate word-level timestamps. We then automatically add punctuation to the transcript via the \verb|deepmultilingualpunctuation| tool~\cite{guhr-EtAl:2021:fullstop}. Finally, we use PySBD~\cite{sadvilkar-neumann-2020-pysbd} to segment the transcript into sentences, such that sentence-wise timestamps can be inferred from the word-level timestamps. We then compute 2\,Hz textual features by averaging the embeddings of all sentences whose timestamps overlap with the respective 500\,ms windows. Since the videos in \hume are typically only a few seconds long, we do not conduct a temporal alignment of their transcripts. The biosignals in \ac{Ulm-TSST} are originally sampled at a 1\,KHz rate. In addition, we downsample them to 2\,Hz, and, subsequently, smooth them via a Savitzky-Golay filter. We make both the 1\,kHz and the downsampled biosignals available. Since \ac{Ulm-TSST} is also labelled at a 2\,Hz rate, the features provided for \ac{MuSe-Personalisation} are already aligned to the annotation signals. As the labels in \ac{Passau-SFCH} refer to windows of size 2\,s, the features for \ac{MuSe-Humor} are not directly aligned to the annotations, but can easily be matched to them. In \ac{MuSe-Mimic}, the labels refer to whole videos, such that no alignment to the labels is necessary.

\subsection{Baseline Training\label{sec:model}}
Given that all tasks are based on video recordings and thus of sequential nature, we opt for a \ac{GRU}-\ac{RNN} followed by two feed-forward layers to serve as a competitive, yet simple baseline system. We conduct a hyperparameter search on this model for each individual sub-challenge and feature. More specifically, we optimise the \ac{GRU}'s hidden representations' size, the number of stacked \ac{GRU} layers and the learning rate. Furthermore, we consider both unidirectional and bidirectional \acp{GRU}. Having trained all unimodal models, we also experiment with a simple late fusion approach to obtain multimodal baselines. More specifically, we average the best unimodal models' predictions, weighted by the models' performance on the development set. We make the code, hyperparameters, and best model checkpoints publicly available on GitHub\footnote{ \href{https://github.com/EIHW/MuSe-2023}{https://github.com/EIHW/MuSe-2023}}. Following, we outline task-specific training details.

\subsubsection{\ac{MuSe-Mimic}}
In order to predict the scores for the three clip-level emotion annotations in \hume, we encode a video by feeding the video's features into the \ac{GRU} model and taking its final hidden representation. The model is trained utilising \ac{MSE} loss. Due to the large size of \hume, we conduct the hyperparameter search on a subset of the data comprising 5\,000 data points. 

\subsubsection{\ac{MuSe-Humor}}
Since \ac{Passau-SFCH}'s labels refer to 2\,s frames while the features are sampled every 500\,ms, one training data point corresponds to a sequence of at most 4 steps. The late fusion is based on the predictions per 2\,s frame. For each hyperparameter configuration, the training routine is run with 5 different, fixed random seeds. Due to the binary prediction target, we aim for the binary cross-entropy loss function.

\subsubsection{\ac{MuSe-Personalisation}}
Aiming to leverage both the full \ac{Ulm-TSST} dataset and the test-subject specific data, we employ a two-stage approach for the \ac{MuSe-Personalisation} baseline. Following \cite{kathan2022personalised}, we first train a model on the whole dataset. Specifically, we do so via a hyperparameter search using 3 fixed random seeds. We then select the single best model checkpoint based on performance on the development set. In the second step, this model is duplicated for every test subject and further trained on the subjects' data only, employing 5 fixed random seeds. For that, another hyperparameter search optimising the number of epochs and the learning rate is conducted. The final predictions are obtained by taking for every test subject the best among the 5 subject-specific models and having it predict the respective subjects' test labels. %

\begin{figure*}\label{fig:pers_model}
    \centering
    \includegraphics[scale=1.]{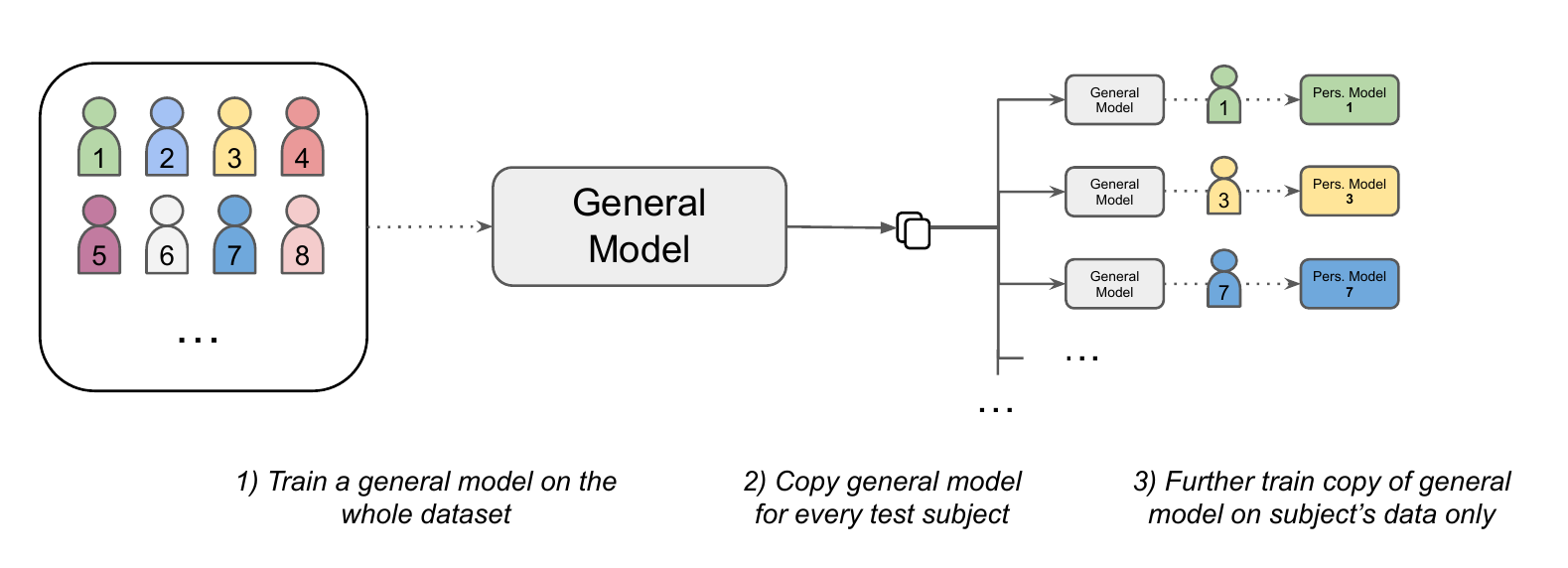}
    \caption{Training method employed in \ac{MuSe-Personalisation}. Note that the IDs in the figure do not correspond to actual IDs in the \ac{Ulm-TSST} dataset.}
    \label{fig:my_label}
\end{figure*}

If, however, the best subject-specific model does not outperform the initial model on the respective subject's development data, we take the initial model's predictions for this subject instead. In both stages, the training data is segmented. Window size and step length of the segmentation are optimised within the hyperparameter search. 
Analogously to previous MuSe challenges~\cite{stappen2020muse1,stappen2021muse, christ2022muse}, we employ \ac{CCC}-Loss as the loss function. We limit our experiments to audio and video-based features and leave experimentation with both physiological signals and textual features to the participants.

\section{Baseline Results}\label{sec:results}

We train \acp{GRU} as described in the previous section. In the following, we present and discuss the baseline results.

\subsection{\ac{MuSe-Mimic}}
The results for MuSe-Mimic are given in Table 3. 

\begin{table}[h!]
\caption{Results for \ac{MuSe-Mimic}. We report the Pearson's correlation coefficient ($\rho$) for the mean of the 3 emotional targets, \textit{Approval, Disappointment, Uncertainty}. The reported score is the best among 5 fixed seeds, as well as the mean $\rho$ over these seeds and the corresponding standard deviations.}

\resizebox{1\columnwidth}{!}{%

\centering
 \begin{tabular}{lcc}
 \toprule 
 & \multicolumn{2}{c}{[Mean $\rho$]} \\
 Features & \multicolumn{1}{c}{Development} & \multicolumn{1}{c}{Test}  \\ \midrule \midrule
 
 \multicolumn{3}{l}{\textbf{Audio}} \\
 \ac{eGeMAPS} & .0842 (.0739 $\pm$ .0067) & .0546 (.0462 $\pm$ .0070) \\
 \ds & .0800 (.0748 $\pm$ .0043) & .0708 (.0734 $\pm$ .0072) \\
  \wtv & .4317 (.4290 $\pm$ .0020) & .4296 (.4330) $\pm$ .0029) \\
 \midrule
 
  \multicolumn{3}{l}{\textbf{Video}} \\
 \ac{FAU} & .1280 (.1241 $\pm$ .0032) & .1337 (.1319 $\pm$ .0019) \\
 \vit & .1202 (.1151 $\pm$ .0041) & .1068 (.1046 $\pm$ .0098) \\
\facenet & .0669 (.0540 $\pm$ .0072) & .0292 (.0275 $\pm$ .0160) \\
 \midrule
 
 \multicolumn{3}{l}{\textbf{Text}} \\
 \textsc{ELECTRA} & .4079 (.4027 $\pm$ .0028) & .3855 (.3902 $\pm$ .0037)
 \\
 \midrule
 \multicolumn{3}{l}{\textbf{Late Fusion}} \\
 A+T & .4718 (.4695 $\pm$ .0022) & .4679 (.4657 $\pm$ .0025) \\
 A+V & .4234 (.4131 $\pm$ .0094) & .4281 (.4209 $\pm$ .0079) \\
 T+V & .4027 (.3983 $\pm$ .0049) & .3965 (.3869 $\pm$ .0075) \\
 A+T+V & .4789 (.4761 $\pm$ .0024) & \textbf{.4727} (.4711 $\pm$ .0023) \\
 \bottomrule
 \end{tabular}\label{tab:mimic}
}

\end{table}
The results of the experiment show that \wtv features perform best among all unimodal approaches, with Mean Pearson's values of $.4317$ and $.4296$ achieved on the development and test sets, respectively. \textsc{ELECTRA} features exhibit comparable performance, yielding Mean Pearson's values of $.4079$ and $.3855$ on the development and test sets, respectively. In contrast, the other five audio- and video-based features display considerably lower performance, with \facenet's result on the test set ($0.0292$) indicating only a very weak correlation. Furthermore, all face-related features produce relatively low Pearson's values, with \acp{FAU} performing the best among them, $.1337$ on the test set. With respect to the audio modality, \wtv features stand out clearly, surpassing the $\rho$ values of both \ac{eGeMAPS} and \ds by a substantial margin. This could be due to the fact that the \wtv model incorporates linguistic knowledge as it is pre-trained on an \ac{ASR} task~\cite{wagner2022dawn}. Moreover, the relatively high Pearson values obtained with textual-only features (\textsc{ELECTRA}) indicate that the textual modality encodes essential information for the task. Therefore, it is likely that the model primarily exploits linguistic information embedded in the \wtv embeddings, thus outperforming the audio-only features \ds and \ac{eGeMAPS}. 

\begin{table}
\caption{Class-wise Pearson's correlation values ($\rho$) for the best feature per modality and the late fusion of all of them. Reported are the mean results across 5 fixed seeds and the respective standard deviations.}\label{tab:classwise}
\resizebox{1\columnwidth}{!}{%

\centering
 \begin{tabular}{lccc}
 \toprule 
 & \multicolumn{3}{c}{[Class-wise $\rho$]} \\
 Features & \multicolumn{1}{c}{Approval} & \multicolumn{1}{c}{Disappointment} & \multicolumn{1}{c}{Uncertainty}  \\ \midrule \midrule

  \wtv (A) 
  & .5139 $\pm$ .0032 
  & .4813 $\pm$ .0023 
  & .3045 $\pm$ .0036 \\
  \textsc{ELECTRA} (T) 
  & .4590 $\pm$ .0055 
  & .4227 $\pm$ .0054 
  & .2889 $\pm$ .0034 \\
 \ac{FAU} (V) 
  & .1786 $\pm$ .0030 
  & .1407 $\pm$ .0041 
  & .0763 $\pm$ .0064 \\

 A+T+V 
  & \textbf{.5536} $\pm$ .0029 
  & \textbf{.5139} $\pm$ .0032 
  & \textbf{.3395} $\pm$ .0015 \\
 \bottomrule
 \end{tabular}
}

\end{table}

\Cref{tab:classwise} provides further insights into the performance of the best models with respect to individual emotion classes. The analysis reveals that, for all three labels, the highest mean class-wise correlation is achieved for Approval, while Uncertainty consistently proves to be the most challenging emotion target. For instance, in the case of \wtv, a $\rho$ value of $.5139$ is observed for Approval, while the corresponding value for Uncertainty is only $.3045$. This trend is also reflected in the late fusion results (\textit{A+T+V}), which outperform all unimodal class-wise results with mean correlations of $.5536$, $.5139$, and $.3395$ for Approval, Disappointment, and Uncertainty, respectively.

Moreover, as evident from \Cref{tab:mimic} and \Cref{tab:classwise}, that performance is enhanced by the multimodal late fusion approach. Notably, the combination of audio and text yields promising results, improving upon the \wtv outcomes by approximately $.04$ on both the development and test sets. Finally, the fusion of predictions obtained with the best features per modality, namely \wtv, \acp{FAU}, and \textsc{ELECTRA}, leads to a Pearson's of $.4789$ and $.4727$ for development and test, respectively. Therefore, although the textual modality seems to play a critical role in the task at hand, multimodal approaches are expected to outperform text-only methods, indicating that all three modalities contain valuable information that can contribute to the task.

\subsection{\ac{MuSe-Humor}}
\Cref{tab:humor} reports the results for \ac{MuSe-Humor}.

\begin{table}[h!]
\caption{\ac{MuSe-Humor} baseline results. Each line refers to experiments conducted with 5 fixed seeds and reports the best AUC-Score among them, together with the mean \ac{AUC}-Scores and their standard deviations across the 5 seeds.}

\resizebox{1\columnwidth}{!}{%

\centering
 \begin{tabular}{lcc}
 \toprule 
 & \multicolumn{2}{c}{[\ac{AUC}]} \\
 Features & \multicolumn{1}{c}{Development} & \multicolumn{1}{c}{Test}  \\ \midrule \midrule
 
 \multicolumn{3}{l}{\textbf{Audio}} \\
 \ac{eGeMAPS} & .7235 (.6836 $\pm$ .0254) & .6672 (.6554 $\pm$ .0169) \\
 \ds & .6969 (.6936 $\pm$ .0022) & .7012 (.7019 $\pm$ .0025) \\
  \wtv & .8435 (.8332 $\pm$ .0082) & .7940 (.7929 $\pm$ .0113) \\
 \midrule
 
  \multicolumn{3}{l}{\textbf{Video}} \\
 \ac{FAU} & .7879 (.7702 $\pm$ .0087) & .6398 (.5948 $\pm$ .0546) \\
 \vit & .8277 (.7890 $\pm$ .0257) & .7457 (.7478 $\pm$ .0093) \\
\facenet & .7342 (.6608 $\pm$ .0671) & .5350 (.5446 $\pm$ .0188) \\
 \midrule
 
 \multicolumn{3}{l}{\textbf{Text}} \\
 \textsc{BERT} & .8105 (.7635 $\pm$ .0717) & .7572 (.7108 $\pm$ .0830)
 \\
 \midrule
 \multicolumn{3}{l}{\textbf{Late Fusion}} \\
 A+T & .8791 (.8600 $\pm$ .0218) & .8218 (.8067 $\pm$ .0149) \\
 A+V & .8656 (.8362 $\pm$ .0205) & .8222 (.8125 $\pm$ .0112) \\
 T+V & .8428 (.8125 $\pm$ .0246) & .7907 (.7780 $\pm$ .0259) \\
 A+T+V & .8759 (.8504 $\pm$ .0209) & \textbf{.8310} (.8244 $\pm$ .0168) \\
 \bottomrule
 \end{tabular}\label{tab:humor}
}

\end{table}

All features lead to results above chance, \ie $0.5$ \ac{AUC}. With \ac{AUC}-Scores of $.8435$ and $.7940$ on the development and test data, respectively, \wtv is the best-performing feature overall. Analogously to the results for \ac{MuSe-Mimic}, one reason might be that \wtv, pretrained on an \ac{ASR} task, also encodes linguistic information already~\cite{wagner2022dawn} and can thus not be regarded as a strictly unimodal feature extractor. It clearly outperforms \ac{eGeMAPS} and \ds, which are, by construction, based on audio signals only. \ac{eGeMAPS} and \ds generalise well to the test data, with the mean \ac{AUC} for \ac{eGeMAPS} decreasing from $.6836$ to $.6554$ and the mean \ac{AUC} of \ds even improving slightly from $.6936$ on the development data to $.7019$ on the test data set. In contrast, notable generalisation issues can be observed for the video modality, in particular for the \ac{FAU} and \facenet features. The mean \ac{AUC}-Score obtained with \ac{FAU} features drops from $.7702$ on the development data to $.5948$ when evaluating on the cross-cultural test data. Considering only the best seed, the textual features ($.7572$ \ac{AUC}) outperform all other features but \wtv on the test set. However, their performance is rather unstable. Their standard deviations of $.0717$ on the development data and $.0830$ on the test data are higher than those of all other features.

As for the simple late fusion approach, all possible modality combinations yield improvements over the respective unimodal results, demonstrating the multimodal nature of humour expression. The combination of the best audio, text, and video models accounts for the best \ac{AUC}-Score on the test set overall, namely $.8310$.

\subsection{\ac{MuSe-Personalisation}}
\Cref{tab:stress} reports the results obtained for \ac{MuSe-Personalisation}. In every experiment besides Valence prediction with \wtv, the result on the test partition is considerably lower than the CCC value on the development set. For both modalities and prediction targets, at least one feature set exists that leads to a \ac{CCC} value larger than 0.5 on the test partition. Regarding the video modality, \facenet clearly trumps the other two feature sets for both arousal and valence, with its arousal \ac{CCC} value of $.5959$ on the test data exceeding the second best video feature's (\acp{FAU}) by more than $0.200$. Moreover, \facenet is the best-performing feature for valence, surpassing all other unimodal results on both the development and the test data. As for audio, there is no such clearly superior feature set. While \ac{eGeMAPS} accounts for the best results on the development data of both arousal and valence, its result on the test sets are outperformed by \ds for arousal and \wtv for valence. In general, audio features can be said to achieve better results for predicting arousal than face-based feature sets. The \ac{CCC} values of both \ac{eGeMAPS} and \ds on the test set for arousal, namely $.7395$ and $.7482$, outperform the best visual-based result, \ie the \ac{CCC} value of $.5959$ obtained using \facenet. For valence, such a tendency does not exist. Here, only one feature set per modality yields a \ac{CCC} value slightly above 0.5 on the test data, namely \wtv accounting for a \ac{CCC} of $.5232$ and \facenet leading to a \ac{CCC} of $.5654$. 

The late fusion results demonstrate the complementarity of the audio and video modality. In particular, the fusion of \ac{eGeMAPS} and \facenet for valence considerably improves upon their individual performance on the test set, resulting in a \ac{CCC} value of $.7827$. This effect is less extreme regarding arousal, where the late fusion result of $.7450$ on the test set only slightly exceeds the respective result achieved utilising \ac{eGeMAPS} only, \ie $.7395$. 
Overall, the best combined score of $.7639$ on the test set is obtained using the late fusion approach. The differences between one feature's performance on the development and on the test data can be large. This issue might be addressed by a more sophisticated usage of the provided data. In particular, participants are not bound to use the first 60 seconds as training and the second minute as development data.
\begin{table}[h!]

\caption{Results for \ac{MuSe-Personalisation}. Reported are the \ac{CCC} values for valence, and physiological arousal after subject-specific training, \ie, based on the best personalised model among five seeds for each subject. \emph{Combined} refers to the mean of the respective feature's valence and arousal \ac{CCC} values. 
}
\resizebox{1.0\columnwidth}{!}{%
\centering
 \begin{tabular}{lccccc}
 \toprule
  & \multicolumn{2}{c}{\textbf{Arousal}} & \multicolumn{2}{c}{\textbf{Valence}} & \textbf{Combined} \\ 
  & \multicolumn{2}{c}{[\ac{CCC}]} & \multicolumn{2}{c}{[\ac{CCC}]} & [\ac{CCC}] \\
 Features   & Dev. & Test   & Dev. & Test & Test \\ 
 \midrule \midrule

 \multicolumn{6}{l}{\textbf{Audio}} \\
  \ac{eGeMAPS}   & .9073 & .7395 & .5892  & .3944 & .5670 \\
  
  \ds   & .8064 & \textbf{.7482} & .3536  & .2836 & .5159 \\ 
  
  \wtv   & .7421 & .5325 & .5142  & .5232 & .5279 \\
  \midrule

  \multicolumn{6}{l}{\textbf{Video}} \\
  \ac{FAU}   & .6382 & .3766 & .1468  & .1076 & .2421 \\
  
  \vit   & .2691 & .0001 & .6050  & .4490 & .2246 \\
  
  \facenet   & .8260 & .5959 & .6491  & .5654 & .5807 \\
  \midrule

  \multicolumn{6}{l}{\textbf{Late Fusion}} \\
A + V & .9145 & .7450 & .8559  & \textbf{.7827} & \textbf{.7639} \\  

 \bottomrule 
 \end{tabular}
}
\label{tab:stress}
\end{table}

\section{Conclusions}\label{sec:conclusion}
We introduced MuSe 2023 -- the 4th Multimodal Sentiment Analysis challenge. 
MuSe 2023 comprises three sub-challenges: 

For the \ac{MuSe-Mimic} Sub-Challenge, the novel \hume data set is made available, consisting of recordings of people mimicking emotional videos. Participants predict the degree of Approval, Disappointment, and Uncertainty for each video. 

The \ac{MuSe-Humor} Sub-Challenge is based on an extension of the \ac{Passau-SFCH} dataset~\cite{christ2022multimodal}, aiming at the recognition of spontaneous humour in press conferences. A cross-cultural setting is put forward for this task, where participants train their model on German recordings, but have to predict humorous utterances in English data. 

In the \ac{MuSe-Personalisation} Sub-Challenge, participants are encouraged to develop methods for tailoring models to specific individuals. \ac{MuSe-Personalisation} employs the \ac{Ulm-TSST} dataset already featured in previous iterations of MuSe~\cite{stappen2021muse, christ2022muse}. While all sub-challenges feature rather simple scenarios involving one person, we believe that the variety of prediction targets and modalities will also foster progress in empathetic dialogue systems and conversational sentiment analysis.

We utilised publicly available software to pre-compute a wide selection of features for the audio, video and text modalities. Moreover, on the basis of these features, we trained a simple \ac{GRU} model to obtain the following official baseline results on the respective test sets: for \ac{MuSe-Mimic}, a mean $\rho$ value of $.4727$ was obtained via the late fusion of the audio, video, and text features. Similarly, for \ac{MuSe-Humor} a late fusion of all three modalities accounts for the baseline \ac{AUC} value of $.8310$. Regarding \ac{MuSe-Personalisation}, \ds features lead to the best arousal \ac{CCC} value of $.7482$, while a late fusion of audio and video features yielded the best results for both valence ($.7827$ \ac{CCC}) and the mean of arousal and valence, namely $.7639$ \ac{CCC}.

We made the code, data sets and features publicly available. 
The results of our fairly simple baseline systems provide first insights into the suitability of the different modalities and features for the proposed sub-challenges. We expect that these baseline results can be improved considerably via more sophisticated models and methods. MuSe 2023 is intended to be a common platform for developing and evaluating such novel multimodal approaches.

For future efforts, beyond the mere optimisation of performances targeted by this challenge, many more will need to be faced, including pressing aspects such as dependability, explainability, fairness, `green' efficient processing, to name but some of the most urgent ones. 
            
\section{Acknowledgments}
This project has received funding from the 
Deutsche Forschungsgemeinschaft (DFG) under grant agreement No.\ 461420398,
and the DFG's Reinhart Koselleck project No.\ 442218748 (AUDI0NOMOUS).

\begin{acronym}
\acro{AReLU}[AReLU]{Attention-based Rectified Linear Unit}
\acro{AUC}[AUC]{Area Under the Curve}
\acro{ASR}[ASR]{Automatic Speech Recognition}
\acro{CCC}[CCC]{Concordance Correlation Coefficient}
\acro{CNN}[CNN]{Convolutional Neural Network}
\acrodefplural{CNN}[CNNs]{Convolutional Neural Networks}
\acro{CI}[CI]{Confidence Interval}
\acrodefplural{CI}[CIs]{Confidence Intervals}
\acro{CCS}[CCS]{COVID-19 Cough}
\acro{CSS}[CSS]{COVID-19 Speech}
\acro{CTW}[CTW]{Canonical Time Warping}
\acro{ComParE}[ComParE]{Computational Paralinguistics Challenge}
\acrodefplural{ComParE}[ComParE]{Computational Paralinguistics Challenges}
\acro{DNN}[DNN]{Deep Neural Network}
\acrodefplural{DNNs}[DNNs]{Deep Neural Networks}
\acro{DEMoS}[DEMoS]{Database of Elicited Mood in Speech}
\acro{eGeMAPS}[\textsc{eGeMAPS}]{extended Geneva Minimalistic Acoustic Parameter Set}
\acro{EULA}[EULA]{End User License Agreement}
\acro{EWE}[EWE]{Evaluator Weighted Estimator}
\acro{FLOP}[FLOP]{Floating Point Operation}
\acrodefplural{FLOP}[FLOPs]{Floating Point Operations}
\acro{FAU}[FAU]{Facial Action Unit}
\acrodefplural{FAU}[FAUs]{Facial Action Units}
\acro{GDPR}[GDPR]{General Data Protection Regulation}
\acro{GRU}[GRU]{Gated Recurrent Unit}
\acro{HDF}[HDF]{Hierarchical Data Format}
\acro{Hume-Reaction}[\textsc{Hume-Reaction}]{Hume-Reaction}
\acro{HSQ}[HSQ]{Humour Style Questionnaire}
\acro{IEMOCAP}[IEMOCAP]{Interactive Emotional Dyadic Motion Capture}
\acro{KSS}[KSS]{Karolinska Sleepiness Scale}
\acro{LIME}[LIME]{Local Interpretable Model-agnostic Explanations}
\acro{LLD}[LLD]{Low-Level Descriptor}
\acrodefplural{LLD}[LLDs]{Low-Level Descriptors}
\acro{LSTM}[LSTM]{Long Short-Term Memory}
\acro{MIP}[MIP]{Mood Induction Procedure}
\acro{MIP}[MIPs]{Mood Induction Procedures}
\acro{MLP}[MLP]{Multilayer Perceptron}
\acrodefplural{MLP}[MLPs]{Multilayer Perceptrons}
\acro{MPSSC}[MPSSC]{Munich-Passau Snore Sound Corpus}
\acro{MSE}[MSE]{Mean Squared Error}
\acro{MTCNN}[MTCNN]{Multi-task Cascaded Convolutional Networks}
\acro{MuSe}[MuSe]{\textbf{Mu}ltimodal \textbf{Se}ntiment Analysis Challenge}
\acro{MuSe-Humor}[\textsc{MuSe-Humour}]{Cross-Cultural Humour Detection Sub-Challenge}
\acro{MuSe-Mimic}[\textsc{MuSe-Mimic}]{Mimicked Emotions Sub-Challenge}
\acro{MuSe-Stress}[\textsc{MuSe-Stress}]{Emotional Stress Sub-Challenge}
\acro{MuSe-Personalisation}[\textsc{MuSe-Personalisation}]{Personalisation Sub-Challenge}
\acro{Passau-SFCH}[\textsc{Passau-SFCH}]{Passau Spontaneous Football Coach Humour}
\acro{RAAW}[\textsc{RAAW}]{Rater Aligned Annotation Weighting}
\acro{RAVDESS}[RAVDESS]{Ryerson Audio-Visual Database of Emotional Speech and Song}
\acro{RNN}[RNN]{Recurrent Neural Network}
\acro{SER}[SER]{Speech Emotion Recognition}
\acro{SHAP}[SHAP]{SHapley Additive exPlanations}
\acro{SLEEP}[SLEEP]{Düsseldorf Sleepy Language Corpus}
\acro{STFT}[STFT]{Short-Time Fourier Transform}
\acrodefplural{STFT}[STFTs]{Short-Time Fourier Transforms}
\acro{SVM}[SVM]{Support Vector Machine}
\acro{TF}[TF]{TensorFlow}
\acro{TSST}[TSST]{Trier Social Stress Test}
\acro{TNR}[TNR]{True Negative Rate}
\acro{TPR}[TPR]{True Positive Rate}
\acro{UAR}[UAR]{Unweighted Average Recall}
\acro{Ulm-TSST}[\textsc{Ulm-TSST}]{Ulm-Trier Social Stress Test}
\acrodefplural{UAR}[UARs]{Unweighted Average Recall}
\end{acronym}

\clearpage
\footnotesize
\bibliographystyle{ACM-Reference-Format}
\balance
\bibliography{sample-base}

%%% -*-BibTeX-*-
%%% Do NOT edit. File created by BibTeX with style
%%% ACM-Reference-Format-Journals [18-Jan-2012].

\begin{thebibliography}{67}

%%% ====================================================================
%%% NOTE TO THE USER: you can override these defaults by providing
%%% customized versions of any of these macros before the \bibliography
%%% command.  Each of them MUST provide its own final punctuation,
%%% except for \shownote{}, \showDOI{}, and \showURL{}.  The latter two
%%% do not use final punctuation, in order to avoid confusing it with
%%% the Web address.
%%%
%%% To suppress output of a particular field, define its macro to expand
%%% to an empty string, or better, \unskip, like this:
%%%
%%% \newcommand{\showDOI}[1]{\unskip}   % LaTeX syntax
%%%
%%% \def \showDOI #1{\unskip}           % plain TeX syntax
%%%
%%% ====================================================================

\ifx \showCODEN    \undefined \def \showCODEN     #1{\unskip}     \fi
\ifx \showDOI      \undefined \def \showDOI       #1{#1}\fi
\ifx \showISBNx    \undefined \def \showISBNx     #1{\unskip}     \fi
\ifx \showISBNxiii \undefined \def \showISBNxiii  #1{\unskip}     \fi
\ifx \showISSN     \undefined \def \showISSN      #1{\unskip}     \fi
\ifx \showLCCN     \undefined \def \showLCCN      #1{\unskip}     \fi
\ifx \shownote     \undefined \def \shownote      #1{#1}          \fi
\ifx \showarticletitle \undefined \def \showarticletitle #1{#1}   \fi
\ifx \showURL      \undefined \def \showURL       {\relax}        \fi
% The following commands are used for tagged output and should be
% invisible to TeX
\providecommand\bibfield[2]{#2}
\providecommand\bibinfo[2]{#2}
\providecommand\natexlab[1]{#1}
\providecommand\showeprint[2][]{arXiv:#2}

\bibitem[Akyunus et~al\mbox{.}(2021)]%
        {akyunus2021age}
\bibfield{author}{\bibinfo{person}{Miray Akyunus}, \bibinfo{person}{T{\"u}lin
  Gen{\c{c}}{\"o}z}, {and} \bibinfo{person}{B~T{\"u}rk{\"u}ler Aka}.}
  \bibinfo{year}{2021}\natexlab{}.
\newblock \showarticletitle{Age and sex differences in basic personality traits
  and interpersonal problems across young adulthood}.
\newblock \bibinfo{journal}{\emph{Current Psychology}}  \bibinfo{volume}{40}
  (\bibinfo{year}{2021}), \bibinfo{pages}{2518--2527}.
\newblock


\bibitem[Amiriparian et~al\mbox{.}(2022a)]%
        {amiriparian2022muse}
\bibfield{author}{\bibinfo{person}{Shahin Amiriparian}, \bibinfo{person}{Lukas
  Christ}, \bibinfo{person}{Andreas K{\"o}nig}, \bibinfo{person}{Eva-Maria
  Me{\ss}ner}, \bibinfo{person}{Alan Cowen}, \bibinfo{person}{Erik Cambria},
  {and} \bibinfo{person}{Bj{\"o}rn~W Schuller}.}
  \bibinfo{year}{2022}\natexlab{a}.
\newblock \showarticletitle{MuSe 2022 Challenge: Multimodal Humour, Emotional
  Reactions, and Stress}. In \bibinfo{booktitle}{\emph{Proceedings of the 30th
  ACM International Conference on Multimedia}}. \bibinfo{publisher}{Association
  for Computing Machinery}, \bibinfo{address}{New York, NY, USA},
  \bibinfo{pages}{7389--7391}.
\newblock


\bibitem[Amiriparian et~al\mbox{.}(2017a)]%
        {Amiriparian17-SAU}
\bibfield{author}{\bibinfo{person}{Shahin Amiriparian},
  \bibinfo{person}{Nicholas Cummins}, \bibinfo{person}{Sandra Ottl},
  \bibinfo{person}{Maurice Gerczuk}, {and} \bibinfo{person}{Bj\"orn Schuller}.}
  \bibinfo{year}{2017}\natexlab{a}.
\newblock \showarticletitle{{Sentiment Analysis Using Image-based Deep Spectrum
  Features}}. In \bibinfo{booktitle}{\emph{{Proceedings 2nd International
  Workshop on Automatic Sentiment Analysis in the Wild (WASA 2017) held in
  conjunction with the 7th biannual Conference on Affective Computing and
  Intelligent Interaction (ACII 2017)}}}. AAAC, \bibinfo{publisher}{IEEE},
  \bibinfo{address}{San Antonio, TX}, \bibinfo{pages}{26--29}.
\newblock


\bibitem[Amiriparian et~al\mbox{.}(2017b)]%
        {Amiriparian17-SSC}
\bibfield{author}{\bibinfo{person}{Shahin Amiriparian},
  \bibinfo{person}{Maurice Gerczuk}, \bibinfo{person}{Sandra Ottl},
  \bibinfo{person}{Nicholas Cummins}, \bibinfo{person}{Michael Freitag},
  \bibinfo{person}{Sergey Pugachevskiy}, {and} \bibinfo{person}{Bj\"orn
  Schuller}.} \bibinfo{year}{2017}\natexlab{b}.
\newblock \showarticletitle{{Snore Sound Classification Using Image-based Deep
  Spectrum Features}}. In \bibinfo{booktitle}{\emph{{Proceedings INTERSPEECH
  2017, 18th Annual Conference of the International Speech Communication
  Association}}}. ISCA, \bibinfo{publisher}{ISCA}, \bibinfo{address}{Stockholm,
  Sweden}, \bibinfo{pages}{3512--3516}.
\newblock


\bibitem[Amiriparian et~al\mbox{.}(2020)]%
        {Amiriparian20-TCP}
\bibfield{author}{\bibinfo{person}{Shahin Amiriparian},
  \bibinfo{person}{Maurice Gerczuk}, \bibinfo{person}{Lukas Stappen},
  \bibinfo{person}{Alice Baird}, \bibinfo{person}{Lukas Koebe},
  \bibinfo{person}{Sandra Ottl}, {and} \bibinfo{person}{Bj\"orn Schuller}.}
  \bibinfo{year}{2020}\natexlab{}.
\newblock \showarticletitle{{Towards Cross-Modal Pre-Training and Learning
  Tempo-Spatial Characteristics for Audio Recognition with Convolutional and
  Recurrent Neural Networks}}.
\newblock \bibinfo{journal}{\emph{EURASIP Journal on Audio, Speech, and Music
  Processing}} \bibinfo{volume}{2020}, \bibinfo{number}{19}
  (\bibinfo{year}{2020}), \bibinfo{pages}{1--11}.
\newblock


\bibitem[Amiriparian et~al\mbox{.}(2022b)]%
        {Amiriparian22-DAP}
\bibfield{author}{\bibinfo{person}{Shahin Amiriparian}, \bibinfo{person}{Tobias
  Hübner}, \bibinfo{person}{Vincent Karas}, \bibinfo{person}{Maurice Gerczuk},
  \bibinfo{person}{Sandra Ottl}, {and} \bibinfo{person}{Björn~W. Schuller}.}
  \bibinfo{year}{2022}\natexlab{b}.
\newblock \showarticletitle{DeepSpectrumLite: A Power-Efficient Transfer
  Learning Framework for Embedded Speech and Audio Processing From
  Decentralized Data}.
\newblock \bibinfo{journal}{\emph{Frontiers in Artificial Intelligence}}
  \bibinfo{volume}{5} (\bibinfo{year}{2022}), \bibinfo{numpages}{10}~pages.
\newblock
\showISSN{2624-8212}
\urldef\tempurl%
\url{https://doi.org/10.3389/frai.2022.856232}
\showDOI{\tempurl}


\bibitem[Baevski et~al\mbox{.}(2020)]%
        {baevski2020wav2vec}
\bibfield{author}{\bibinfo{person}{Alexei Baevski}, \bibinfo{person}{Yuhao
  Zhou}, \bibinfo{person}{Abdelrahman Mohamed}, {and} \bibinfo{person}{Michael
  Auli}.} \bibinfo{year}{2020}\natexlab{}.
\newblock \showarticletitle{wav2vec 2.0: A framework for self-supervised
  learning of speech representations}.
\newblock \bibinfo{journal}{\emph{Advances in neural information processing
  systems}}  \bibinfo{volume}{33} (\bibinfo{year}{2020}),
  \bibinfo{pages}{12449--12460}.
\newblock


\bibitem[Baird et~al\mbox{.}(2019)]%
        {baird2019can}
\bibfield{author}{\bibinfo{person}{Alice Baird}, \bibinfo{person}{Shahin
  Amiriparian}, {and} \bibinfo{person}{Bj{\"o}rn Schuller}.}
  \bibinfo{year}{2019}\natexlab{}.
\newblock \showarticletitle{Can deep generative audio be emotional? Towards an
  approach for personalised emotional audio generation}. In
  \bibinfo{booktitle}{\emph{2019 IEEE 21st International Workshop on Multimedia
  Signal Processing (MMSP)}}. IEEE, \bibinfo{publisher}{IEEE},
  \bibinfo{address}{Kuala Lumpur, Malaysia}, \bibinfo{pages}{1--5}.
\newblock


\bibitem[Baird et~al\mbox{.}(2021)]%
        {baird2021physiologically}
\bibfield{author}{\bibinfo{person}{Alice Baird}, \bibinfo{person}{Lukas
  Stappen}, \bibinfo{person}{Lukas Christ}, \bibinfo{person}{Lea Schumann},
  \bibinfo{person}{Eva-Maria Me{\ss}ner}, {and} \bibinfo{person}{Bj{\"o}rn~W
  Schuller}.} \bibinfo{year}{2021}\natexlab{}.
\newblock \showarticletitle{A Physiologically-adapted Gold Standard for Arousal
  During a Stress Induced Scenario}. In \bibinfo{booktitle}{\emph{Proceedings
  of the 2nd Multimodal Sentiment Analysis Challenge, co-located with the 29th
  ACM International Conference on Multimedia (ACMMM)}}. ACM,
  \bibinfo{publisher}{Association for Computing Machinery},
  \bibinfo{address}{Changu, China}, \bibinfo{pages}{69–73}.
\newblock


\bibitem[Binsted et~al\mbox{.}(1995)]%
        {binsted1995using}
\bibfield{author}{\bibinfo{person}{Kim Binsted} {et~al\mbox{.}}}
  \bibinfo{year}{1995}\natexlab{}.
\newblock \showarticletitle{Using humour to make natural language interfaces
  more friendly}. In \bibinfo{booktitle}{\emph{Proceedings of the ai, alife and
  entertainment workshop, intern. Joint conf. On artificial intelligence}}.
\newblock


\bibitem[{Bj\"orn W.\ Schuller and Anton Batliner and Christian Bergler and
  Cecilia Mascolo and Jing Han and Iulia Lefter and Heysem Kaya and Shahin
  Amiriparian and Alice Baird and Lukas Stappen and Sandra Ottl and Maurice
  Gerczuk and Panaguiotis Tzirakis and Chlo\"e Brown and Jagmohan Chauhan and
  Andreas Grammenos and Apinan Hasthanasombat and Dimitris Spathis and Tong Xia
  and Pietro Cicuta and Leon J.\,M.\ Rothkrantz and Joeri Zwerts and Jelle
  Treep and Casper Kaandorp}(2021)]%
        {Schuller21-TI2}
\bibfield{author}{\bibinfo{person}{{Bj\"orn W.\ Schuller and Anton Batliner and
  Christian Bergler and Cecilia Mascolo and Jing Han and Iulia Lefter and
  Heysem Kaya and Shahin Amiriparian and Alice Baird and Lukas Stappen and
  Sandra Ottl and Maurice Gerczuk and Panaguiotis Tzirakis and Chlo\"e Brown
  and Jagmohan Chauhan and Andreas Grammenos and Apinan Hasthanasombat and
  Dimitris Spathis and Tong Xia and Pietro Cicuta and Leon J.\,M.\ Rothkrantz
  and Joeri Zwerts and Jelle Treep and Casper Kaandorp}}.}
  \bibinfo{year}{2021}\natexlab{}.
\newblock \showarticletitle{{The INTERSPEECH 2021 Computational Paralinguistics
  Challenge: COVID-19 Cough, COVID-19 Speech, Escalation \& Primates}}. In
  \bibinfo{booktitle}{\emph{{Proceedings INTERSPEECH 2021, 22nd Annual
  Conference of the International Speech Communication Association}}}. ISCA,
  \bibinfo{publisher}{ISCA}, \bibinfo{address}{Brno, Czechia},
  \bibinfo{pages}{431--435}.
\newblock


\bibitem[Brooks et~al\mbox{.}(2023)]%
        {brooks2023deep}
\bibfield{author}{\bibinfo{person}{Jeffrey~A Brooks},
  \bibinfo{person}{Panagiotis Tzirakis}, \bibinfo{person}{Alice Baird},
  \bibinfo{person}{Lauren Kim}, \bibinfo{person}{Michael Opara},
  \bibinfo{person}{Xia Fang}, \bibinfo{person}{Dacher Keltner},
  \bibinfo{person}{Maria Monroy}, \bibinfo{person}{Rebecca Corona},
  \bibinfo{person}{Jacob Metrick}, {et~al\mbox{.}}}
  \bibinfo{year}{2023}\natexlab{}.
\newblock \showarticletitle{Deep learning reveals what vocal bursts express in
  different cultures}.
\newblock \bibinfo{journal}{\emph{Nature Human Behaviour}} \bibinfo{volume}{7},
  \bibinfo{number}{2} (\bibinfo{year}{2023}), \bibinfo{pages}{240--250}.
\newblock


\bibitem[Cann et~al\mbox{.}(2014)]%
        {cann2014assessing}
\bibfield{author}{\bibinfo{person}{Arnie Cann}, \bibinfo{person}{Amanda~J
  Watson}, {and} \bibinfo{person}{Elisabeth~A Bridgewater}.}
  \bibinfo{year}{2014}\natexlab{}.
\newblock \showarticletitle{Assessing humor at work: The humor climate
  questionnaire}.
\newblock \bibinfo{journal}{\emph{Humor}} \bibinfo{volume}{27},
  \bibinfo{number}{2} (\bibinfo{year}{2014}), \bibinfo{pages}{307--323}.
\newblock


\bibitem[Caron et~al\mbox{.}(2021)]%
        {caron2021emerging}
\bibfield{author}{\bibinfo{person}{Mathilde Caron}, \bibinfo{person}{Hugo
  Touvron}, \bibinfo{person}{Ishan Misra}, \bibinfo{person}{Herv{\'e}
  J{\'e}gou}, \bibinfo{person}{Julien Mairal}, \bibinfo{person}{Piotr
  Bojanowski}, {and} \bibinfo{person}{Armand Joulin}.}
  \bibinfo{year}{2021}\natexlab{}.
\newblock \showarticletitle{Emerging properties in self-supervised vision
  transformers}. In \bibinfo{booktitle}{\emph{Proceedings of the IEEE/CVF
  international conference on computer vision}}. \bibinfo{pages}{9650--9660}.
\newblock


\bibitem[Caruelle et~al\mbox{.}(2019)]%
        {caruelle2019use}
\bibfield{author}{\bibinfo{person}{Delphine Caruelle}, \bibinfo{person}{Anders
  Gustafsson}, \bibinfo{person}{Poja Shams}, {and} \bibinfo{person}{Line
  Lervik-Olsen}.} \bibinfo{year}{2019}\natexlab{}.
\newblock \showarticletitle{The use of electrodermal activity (EDA) measurement
  to understand consumer emotions--a literature review and a call for action}.
\newblock \bibinfo{journal}{\emph{Journal of Business Research}}
  \bibinfo{volume}{104} (\bibinfo{year}{2019}), \bibinfo{pages}{146--160}.
\newblock


\bibitem[Chaudhari et~al\mbox{.}(2022)]%
        {chaudhari2022vitfer}
\bibfield{author}{\bibinfo{person}{Aayushi Chaudhari}, \bibinfo{person}{Chintan
  Bhatt}, \bibinfo{person}{Achyut Krishna}, {and} \bibinfo{person}{Pier~Luigi
  Mazzeo}.} \bibinfo{year}{2022}\natexlab{}.
\newblock \showarticletitle{ViTFER: Facial Emotion Recognition with Vision
  Transformers}.
\newblock \bibinfo{journal}{\emph{Applied System Innovation}}
  \bibinfo{volume}{5}, \bibinfo{number}{4} (\bibinfo{year}{2022}),
  \bibinfo{pages}{80}.
\newblock


\bibitem[Chen and Zhang(2022)]%
        {chen2022integrating}
\bibfield{author}{\bibinfo{person}{Chengxin Chen} {and}
  \bibinfo{person}{Pengyuan Zhang}.} \bibinfo{year}{2022}\natexlab{}.
\newblock \showarticletitle{Integrating Cross-Modal Interactions via Latent
  Representation Shift for Multi-Modal Humor Detection}. In
  \bibinfo{booktitle}{\emph{Proceedings of the 3rd International on Multimodal
  Sentiment Analysis Workshop and Challenge}} (Lisboa, Portugal)
  \emph{(\bibinfo{series}{MuSe' 22})}. \bibinfo{publisher}{Association for
  Computing Machinery}, \bibinfo{address}{New York, NY, USA},
  \bibinfo{pages}{23–28}.
\newblock
\showISBNx{9781450394840}
\urldef\tempurl%
\url{https://doi.org/10.1145/3551876.3554805}
\showDOI{\tempurl}


\bibitem[Christ et~al\mbox{.}(2022)]%
        {christ2022multimodal}
\bibfield{author}{\bibinfo{person}{Lukas Christ}, \bibinfo{person}{Shahin
  Amiriparian}, \bibinfo{person}{Alexander Kathan}, \bibinfo{person}{Niklas
  M{\"u}ller}, \bibinfo{person}{Andreas K{\"o}nig}, {and}
  \bibinfo{person}{Bj{\"o}rn~W Schuller}.} \bibinfo{year}{2022}\natexlab{}.
\newblock \showarticletitle{Multimodal Prediction of Spontaneous Humour: A
  Novel Dataset and First Results}.
\newblock \bibinfo{journal}{\emph{arXiv preprint arXiv:2209.14272}}
  (\bibinfo{year}{2022}).
\newblock


\bibitem[{Christ, Lukas and Amiriparian, Shahin and Baird, Alice and Tzirakis,
  Panagiotis and Kathan, Alexander and M\"{u}ller, Niklas and Stappen, Lukas
  and Me\ss{}ner, Eva-Maria and K\"{o}nig, Andreas and Cowen, Alan and Cambria,
  Erik and Schuller, Bj\"{o}rn W.}(2022)]%
        {christ2022muse}
\bibfield{author}{\bibinfo{person}{{Christ, Lukas and Amiriparian, Shahin and
  Baird, Alice and Tzirakis, Panagiotis and Kathan, Alexander and M\"{u}ller,
  Niklas and Stappen, Lukas and Me\ss{}ner, Eva-Maria and K\"{o}nig, Andreas
  and Cowen, Alan and Cambria, Erik and Schuller, Bj\"{o}rn W.}}}
  \bibinfo{year}{2022}\natexlab{}.
\newblock \showarticletitle{{The MuSe 2022 Multimodal Sentiment Analysis
  Challenge: Humor, Emotional Reactions, and Stress}}. In
  \bibinfo{booktitle}{\emph{Proceedings of the 3rd International on Multimodal
  Sentiment Analysis Workshop and Challenge}} (Lisboa, Portugal).
  \bibinfo{publisher}{Association for Computing Machinery},
  \bibinfo{address}{New York, NY, USA}, \bibinfo{pages}{5–14}.
\newblock


\bibitem[Clark et~al\mbox{.}(2020)]%
        {clark2020electra}
\bibfield{author}{\bibinfo{person}{Kevin Clark}, \bibinfo{person}{Minh-Thang
  Luong}, \bibinfo{person}{Quoc~V. Le}, {and} \bibinfo{person}{Christopher~D.
  Manning}.} \bibinfo{year}{2020}\natexlab{}.
\newblock \showarticletitle{{ELECTRA}: Pre-training Text Encoders as
  Discriminators Rather Than Generators}. In \bibinfo{booktitle}{\emph{ICLR}}.
\newblock
\urldef\tempurl%
\url{https://openreview.net/pdf?id=r1xMH1BtvB}
\showURL{%
\tempurl}


\bibitem[Devlin et~al\mbox{.}(2019)]%
        {devlin-etal-2019-bert}
\bibfield{author}{\bibinfo{person}{Jacob Devlin}, \bibinfo{person}{Ming-Wei
  Chang}, \bibinfo{person}{Kenton Lee}, {and} \bibinfo{person}{Kristina
  Toutanova}.} \bibinfo{year}{2019}\natexlab{}.
\newblock \showarticletitle{{BERT}: Pre-training of Deep Bidirectional
  Transformers for Language Understanding}. In
  \bibinfo{booktitle}{\emph{Proceedings of the 2019 Conference of the North
  {A}merican Chapter of the Association for Computational Linguistics: Human
  Language Technologies}}. \bibinfo{pages}{4171--4186}.
\newblock


\bibitem[Doddington et~al\mbox{.}(1998)]%
        {doddington1998sheep}
\bibfield{author}{\bibinfo{person}{George Doddington}, \bibinfo{person}{Walter
  Liggett}, \bibinfo{person}{Alvin Martin}, \bibinfo{person}{Mark Przybocki},
  {and} \bibinfo{person}{Douglas Reynolds}.} \bibinfo{year}{1998}\natexlab{}.
\newblock \bibinfo{booktitle}{\emph{Sheep, goats, lambs and wolves: A
  statistical analysis of speaker performance in the NIST 1998 speaker
  recognition evaluation}}.
\newblock \bibinfo{type}{{T}echnical {R}eport}. \bibinfo{institution}{National
  Inst of Standards and Technology Gaithersburg Md}.
\newblock


\bibitem[Ekman and Friesen(1978)]%
        {ekman1978facial}
\bibfield{author}{\bibinfo{person}{Paul Ekman} {and} \bibinfo{person}{Wallace~V
  Friesen}.} \bibinfo{year}{1978}\natexlab{}.
\newblock \showarticletitle{Facial action coding system}.
\newblock \bibinfo{journal}{\emph{Environmental Psychology \& Nonverbal
  Behavior}} (\bibinfo{year}{1978}).
\newblock


\bibitem[Eyben et~al\mbox{.}(2015)]%
        {eyben2015geneva}
\bibfield{author}{\bibinfo{person}{Florian Eyben}, \bibinfo{person}{Klaus~R
  Scherer}, \bibinfo{person}{Bj{\"o}rn~W Schuller}, \bibinfo{person}{Johan
  Sundberg}, \bibinfo{person}{Elisabeth Andr{\'e}}, \bibinfo{person}{Carlos
  Busso}, \bibinfo{person}{Laurence~Y Devillers}, \bibinfo{person}{Julien
  Epps}, \bibinfo{person}{Petri Laukka}, \bibinfo{person}{Shrikanth~S
  Narayanan}, {et~al\mbox{.}}} \bibinfo{year}{2015}\natexlab{}.
\newblock \showarticletitle{The Geneva minimalistic acoustic parameter set
  (GeMAPS) for voice research and affective computing}.
\newblock \bibinfo{journal}{\emph{IEEE Transactions on Affective Computing}}
  \bibinfo{volume}{7}, \bibinfo{number}{2} (\bibinfo{year}{2015}),
  \bibinfo{pages}{190--202}.
\newblock


\bibitem[Eyben et~al\mbox{.}(2010)]%
        {eyben2010opensmile}
\bibfield{author}{\bibinfo{person}{Florian Eyben}, \bibinfo{person}{Martin
  W{\"o}llmer}, {and} \bibinfo{person}{Bj{\"o}rn Schuller}.}
  \bibinfo{year}{2010}\natexlab{}.
\newblock \showarticletitle{Opensmile: the munich versatile and fast
  open-source audio feature extractor}. In
  \bibinfo{booktitle}{\emph{Proceedings of the 18th ACM International
  Conference on Multimedia}}. \bibinfo{publisher}{Association for Computing
  Machinery}, \bibinfo{address}{Firenze, Italy}, \bibinfo{pages}{1459--1462}.
\newblock


\bibitem[Gerczuk et~al\mbox{.}(2022)]%
        {Gerczuk22-EAT}
\bibfield{author}{\bibinfo{person}{Maurice Gerczuk}, \bibinfo{person}{Shahin
  Amiriparian}, \bibinfo{person}{Sandra Ottl}, {and} \bibinfo{person}{Bj\"orn
  Schuller}.} \bibinfo{year}{2022}\natexlab{}.
\newblock \showarticletitle{{EmoNet: A Transfer Learning Framework for
  Multi-Corpus Speech Emotion Recognition}}.
\newblock \bibinfo{journal}{\emph{IEEE Transactions on Affective Computing}}
  \bibinfo{volume}{13} (\bibinfo{year}{2022}).
\newblock


\bibitem[Gkorezis et~al\mbox{.}(2014)]%
        {gkorezis2014leader}
\bibfield{author}{\bibinfo{person}{Panagiotis Gkorezis},
  \bibinfo{person}{Eugenia Petridou}, {and} \bibinfo{person}{Panteleimon
  Xanthiakos}.} \bibinfo{year}{2014}\natexlab{}.
\newblock \showarticletitle{Leader positive humor and organizational cynicism:
  LMX as a mediator}.
\newblock \bibinfo{journal}{\emph{Leadership \& Organization Development
  Journal}}  \bibinfo{volume}{35} (\bibinfo{year}{2014}), \bibinfo{pages}{305
  -- 315}.
\newblock


\bibitem[Grimm and Kroschel(2005)]%
        {grimm2005evaluation}
\bibfield{author}{\bibinfo{person}{Michael Grimm} {and}
  \bibinfo{person}{Kristian Kroschel}.} \bibinfo{year}{2005}\natexlab{}.
\newblock \showarticletitle{Evaluation of natural emotions using self
  assessment manikins}. In \bibinfo{booktitle}{\emph{IEEE Workshop on Automatic
  Speech Recognition and Understanding, 2005.}} IEEE,
  \bibinfo{publisher}{IEEE}, \bibinfo{address}{Cancún, Mexico},
  \bibinfo{pages}{381--385}.
\newblock


\bibitem[Guhr et~al\mbox{.}(2021)]%
        {guhr-EtAl:2021:fullstop}
\bibfield{author}{\bibinfo{person}{Oliver Guhr}, \bibinfo{person}{Anne-Kathrin
  Schumann}, \bibinfo{person}{Frank Bahrmann}, {and}
  \bibinfo{person}{Hans~Joachim Böhme}.} \bibinfo{year}{2021}\natexlab{}.
\newblock \showarticletitle{FullStop: Multilingual Deep Models for Punctuation
  Prediction}. In \bibinfo{booktitle}{\emph{Proceedings of the Swiss Text
  Analytics Conference 2021}}. \bibinfo{publisher}{CEUR Workshop Proceedings},
  \bibinfo{address}{Winterthur, Switzerland}.
\newblock
\urldef\tempurl%
\url{http://ceur-ws.org/Vol-2957/sepp_paper4.pdf}
\showURL{%
\tempurl}


\bibitem[Hasan et~al\mbox{.}(2019)]%
        {hasan2019ur}
\bibfield{author}{\bibinfo{person}{Md~Kamrul Hasan}, \bibinfo{person}{Wasifur
  Rahman}, \bibinfo{person}{AmirAli Bagher~Zadeh}, \bibinfo{person}{Jianyuan
  Zhong}, \bibinfo{person}{Md~Iftekhar Tanveer},
  \bibinfo{person}{Louis-Philippe Morency}, {and}
  \bibinfo{person}{Mohammed~(Ehsan) Hoque}.} \bibinfo{year}{2019}\natexlab{}.
\newblock \showarticletitle{{UR}-{FUNNY}: A Multimodal Language Dataset for
  Understanding Humor}. In \bibinfo{booktitle}{\emph{Proceedings of the 2019
  Conference on Empirical Methods in Natural Language Processing and the 9th
  International Joint Conference on Natural Language Processing
  (EMNLP-IJCNLP)}}. \bibinfo{publisher}{Association for Computational
  Linguistics}, \bibinfo{address}{Hong Kong, China},
  \bibinfo{pages}{2046--2056}.
\newblock
\urldef\tempurl%
\url{https://doi.org/10.18653/v1/D19-1211}
\showDOI{\tempurl}


\bibitem[He et~al\mbox{.}(2022)]%
        {he2022multimodal}
\bibfield{author}{\bibinfo{person}{Yu He}, \bibinfo{person}{Licai Sun},
  \bibinfo{person}{Zheng Lian}, \bibinfo{person}{Bin Liu},
  \bibinfo{person}{Jianhua Tao}, \bibinfo{person}{Meng Wang}, {and}
  \bibinfo{person}{Yuan Cheng}.} \bibinfo{year}{2022}\natexlab{}.
\newblock \showarticletitle{Multimodal Temporal Attention in Sentiment
  Analysis}. In \bibinfo{booktitle}{\emph{Proceedings of the 3rd International
  on Multimodal Sentiment Analysis Workshop and Challenge}} (Lisboa, Portugal)
  \emph{(\bibinfo{series}{MuSe' 22})}. \bibinfo{publisher}{Association for
  Computing Machinery}, \bibinfo{address}{New York, NY, USA},
  \bibinfo{pages}{61–66}.
\newblock
\showISBNx{9781450394840}
\urldef\tempurl%
\url{https://doi.org/10.1145/3551876.3554811}
\showDOI{\tempurl}


\bibitem[Hess and Fischer(2013)]%
        {hess2013emotional}
\bibfield{author}{\bibinfo{person}{Ursula Hess} {and} \bibinfo{person}{Agneta
  Fischer}.} \bibinfo{year}{2013}\natexlab{}.
\newblock \showarticletitle{Emotional mimicry as social regulation}.
\newblock \bibinfo{journal}{\emph{Personality and social psychology review}}
  \bibinfo{volume}{17}, \bibinfo{number}{2} (\bibinfo{year}{2013}),
  \bibinfo{pages}{142--157}.
\newblock


\bibitem[Hess and Fischer(2014)]%
        {hess2014emotional}
\bibfield{author}{\bibinfo{person}{Ursula Hess} {and} \bibinfo{person}{Agneta
  Fischer}.} \bibinfo{year}{2014}\natexlab{}.
\newblock \showarticletitle{Emotional mimicry: Why and when we mimic emotions}.
\newblock \bibinfo{journal}{\emph{Social and personality psychology compass}}
  \bibinfo{volume}{8}, \bibinfo{number}{2} (\bibinfo{year}{2014}),
  \bibinfo{pages}{45--57}.
\newblock


\bibitem[Kathan et~al\mbox{.}(2022)]%
        {kathan2022personalised}
\bibfield{author}{\bibinfo{person}{Alexander Kathan}, \bibinfo{person}{Shahin
  Amiriparian}, \bibinfo{person}{Lukas Christ}, \bibinfo{person}{Andreas
  Triantafyllopoulos}, \bibinfo{person}{Niklas M\"{u}ller},
  \bibinfo{person}{Andreas K\"{o}nig}, {and} \bibinfo{person}{Bj\"{o}rn~W.
  Schuller}.} \bibinfo{year}{2022}\natexlab{}.
\newblock \showarticletitle{A Personalised Approach to Audiovisual Humour
  Recognition and Its Individual-Level Fairness}. In
  \bibinfo{booktitle}{\emph{Proceedings of the 3rd International on Multimodal
  Sentiment Analysis Workshop and Challenge}} (Lisboa, Portugal)
  \emph{(\bibinfo{series}{MuSe' 22})}. \bibinfo{publisher}{Association for
  Computing Machinery}, \bibinfo{address}{New York, NY, USA},
  \bibinfo{pages}{29–36}.
\newblock
\showISBNx{9781450394840}
\urldef\tempurl%
\url{https://doi.org/10.1145/3551876.3554800}
\showDOI{\tempurl}


\bibitem[Kirschbaum et~al\mbox{.}(1993)]%
        {kirschbaum1993trier}
\bibfield{author}{\bibinfo{person}{Clemens Kirschbaum},
  \bibinfo{person}{Karl-Martin Pirke}, {and} \bibinfo{person}{Dirk~H
  Hellhammer}.} \bibinfo{year}{1993}\natexlab{}.
\newblock \showarticletitle{The ‘Trier Social Stress Test’--a tool for
  investigating psychobiological stress responses in a laboratory setting}.
\newblock \bibinfo{journal}{\emph{Neuropsychobiology}} \bibinfo{volume}{28},
  \bibinfo{number}{1-2} (\bibinfo{year}{1993}), \bibinfo{pages}{76--81}.
\newblock


\bibitem[Ladilova and Schr{\"o}der(2022)]%
        {ladilova2022humor}
\bibfield{author}{\bibinfo{person}{Anna Ladilova} {and} \bibinfo{person}{Ulrike
  Schr{\"o}der}.} \bibinfo{year}{2022}\natexlab{}.
\newblock \showarticletitle{Humor in intercultural interaction: A source for
  misunderstanding or a common ground builder? A multimodal analysis}.
\newblock \bibinfo{journal}{\emph{Intercultural Pragmatics}}
  \bibinfo{volume}{19}, \bibinfo{number}{1} (\bibinfo{year}{2022}),
  \bibinfo{pages}{71--101}.
\newblock


\bibitem[Li et~al\mbox{.}(2023)]%
        {li2023survey}
\bibfield{author}{\bibinfo{person}{Jialin Li}, \bibinfo{person}{Alia Waleed},
  {and} \bibinfo{person}{Hanan Salam}.} \bibinfo{year}{2023}\natexlab{}.
\newblock \showarticletitle{A Survey on Personalized Affective Computing in
  Human-Machine Interaction}.
\newblock \bibinfo{journal}{\emph{arXiv preprint arXiv:2304.00377}}
  (\bibinfo{year}{2023}).
\newblock


\bibitem[Li et~al\mbox{.}(2022)]%
        {li2022hybrid}
\bibfield{author}{\bibinfo{person}{Jia Li}, \bibinfo{person}{Ziyang Zhang},
  \bibinfo{person}{Junjie Lang}, \bibinfo{person}{Yueqi Jiang},
  \bibinfo{person}{Liuwei An}, \bibinfo{person}{Peng Zou},
  \bibinfo{person}{Yangyang Xu}, \bibinfo{person}{Sheng Gao},
  \bibinfo{person}{Jie Lin}, \bibinfo{person}{Chunxiao Fan},
  \bibinfo{person}{Xiao Sun}, {and} \bibinfo{person}{Meng Wang}.}
  \bibinfo{year}{2022}\natexlab{}.
\newblock \showarticletitle{Hybrid Multimodal Feature Extraction, Mining and
  Fusion for Sentiment Analysis}. In \bibinfo{booktitle}{\emph{Proceedings of
  the 3rd International on Multimodal Sentiment Analysis Workshop and
  Challenge}} (Lisboa, Portugal) \emph{(\bibinfo{series}{MuSe' 22})}.
  \bibinfo{publisher}{Association for Computing Machinery},
  \bibinfo{address}{New York, NY, USA}, \bibinfo{pages}{81–88}.
\newblock
\showISBNx{9781450394840}
\urldef\tempurl%
\url{https://doi.org/10.1145/3551876.3554809}
\showDOI{\tempurl}


\bibitem[Liu et~al\mbox{.}(2022a)]%
        {ssl-survey}
\bibfield{author}{\bibinfo{person}{Shuo Liu}, \bibinfo{person}{Adria
  Mallol-Ragolta}, \bibinfo{person}{Emilia Parada-Cabaleiro},
  \bibinfo{person}{Kun Qian}, \bibinfo{person}{Xin Jing},
  \bibinfo{person}{Alexander Kathan}, \bibinfo{person}{Bin Hu}, {and}
  \bibinfo{person}{Björn~W. Schuller}.} \bibinfo{year}{2022}\natexlab{a}.
\newblock \showarticletitle{Audio self-supervised learning: A survey}.
\newblock \bibinfo{journal}{\emph{Patterns}} \bibinfo{volume}{3},
  \bibinfo{number}{12} (\bibinfo{year}{2022}), \bibinfo{pages}{100616}.
\newblock
\showISSN{2666-3899}
\urldef\tempurl%
\url{https://doi.org/10.1016/j.patter.2022.100616}
\showDOI{\tempurl}


\bibitem[Liu et~al\mbox{.}(2022b)]%
        {liu2022improving}
\bibfield{author}{\bibinfo{person}{Yiping Liu}, \bibinfo{person}{Wei Sun},
  \bibinfo{person}{Xing Zhang}, {and} \bibinfo{person}{Yebao Qin}.}
  \bibinfo{year}{2022}\natexlab{b}.
\newblock \showarticletitle{Improving Dimensional Emotion Recognition via
  Feature-Wise Fusion}. In \bibinfo{booktitle}{\emph{Proceedings of the 3rd
  International on Multimodal Sentiment Analysis Workshop and Challenge}}
  (Lisboa, Portugal) \emph{(\bibinfo{series}{MuSe' 22})}.
  \bibinfo{publisher}{Association for Computing Machinery},
  \bibinfo{address}{New York, NY, USA}, \bibinfo{pages}{55–60}.
\newblock
\showISBNx{9781450394840}
\urldef\tempurl%
\url{https://doi.org/10.1145/3551876.3554804}
\showDOI{\tempurl}


\bibitem[Lotfian and Busso(2019)]%
        {Lotfian_2019_3}
\bibfield{author}{\bibinfo{person}{R. Lotfian} {and} \bibinfo{person}{C.
  Busso}.} \bibinfo{year}{2019}\natexlab{}.
\newblock \showarticletitle{Building Naturalistic Emotionally Balanced Speech
  Corpus by Retrieving Emotional Speech From Existing Podcast Recordings}.
\newblock \bibinfo{journal}{\emph{IEEE Transactions on Affective Computing}}
  \bibinfo{volume}{10}, \bibinfo{number}{4} (\bibinfo{date}{October-December}
  \bibinfo{year}{2019}), \bibinfo{pages}{471--483}.
\newblock
\urldef\tempurl%
\url{https://doi.org/10.1109/TAFFC.2017.2736999}
\showDOI{\tempurl}


\bibitem[Martin et~al\mbox{.}(2003)]%
        {martin2003individual}
\bibfield{author}{\bibinfo{person}{Rod~A Martin}, \bibinfo{person}{Patricia
  Puhlik-Doris}, \bibinfo{person}{Gwen Larsen}, \bibinfo{person}{Jeanette
  Gray}, {and} \bibinfo{person}{Kelly Weir}.} \bibinfo{year}{2003}\natexlab{}.
\newblock \showarticletitle{Individual differences in uses of humor and their
  relation to psychological well-being: Development of the Humor Styles
  Questionnaire}.
\newblock \bibinfo{journal}{\emph{Journal of research in personality}}
  \bibinfo{volume}{37}, \bibinfo{number}{1} (\bibinfo{year}{2003}),
  \bibinfo{pages}{48--75}.
\newblock


\bibitem[McAuliffe et~al\mbox{.}(2017)]%
        {mcauliffe2017montreal}
\bibfield{author}{\bibinfo{person}{Michael McAuliffe},
  \bibinfo{person}{Michaela Socolof}, \bibinfo{person}{Sarah Mihuc},
  \bibinfo{person}{Michael Wagner}, {and} \bibinfo{person}{Morgan
  Sonderegger}.} \bibinfo{year}{2017}\natexlab{}.
\newblock \showarticletitle{Montreal Forced Aligner: Trainable Text-Speech
  Alignment Using Kaldi.}. In \bibinfo{booktitle}{\emph{Proceedings of
  INTERSPEECH}}, Vol.~\bibinfo{volume}{2017}. \bibinfo{publisher}{International
  Speech Communication Association (ISCA)}, \bibinfo{address}{Stockholm,
  Sweden}, \bibinfo{pages}{498--502}.
\newblock


\bibitem[Mittal et~al\mbox{.}(2021)]%
        {mittal2021so}
\bibfield{author}{\bibinfo{person}{Anirudh Mittal}, \bibinfo{person}{Pranav
  Jeevan}, \bibinfo{person}{Prerak Gandhi}, \bibinfo{person}{Diptesh Kanojia},
  {and} \bibinfo{person}{Pushpak Bhattacharyya}.}
  \bibinfo{year}{2021}\natexlab{}.
\newblock \bibinfo{title}{" So You Think You're Funny?": Rating the Humour
  Quotient in Standup Comedy}.
\newblock
\newblock
\showeprint{arXiv preprint arXiv:2110.12765}


\bibitem[Morais et~al\mbox{.}(2022)]%
        {morais2022speech}
\bibfield{author}{\bibinfo{person}{Edmilson Morais}, \bibinfo{person}{Ron
  Hoory}, \bibinfo{person}{Weizhong Zhu}, \bibinfo{person}{Itai Gat},
  \bibinfo{person}{Matheus Damasceno}, {and} \bibinfo{person}{Hagai
  Aronowitz}.} \bibinfo{year}{2022}\natexlab{}.
\newblock \showarticletitle{Speech emotion recognition using self-supervised
  features}. In \bibinfo{booktitle}{\emph{ICASSP 2022-2022 IEEE International
  Conference on Acoustics, Speech and Signal Processing (ICASSP)}}. IEEE,
  \bibinfo{pages}{6922--6926}.
\newblock


\bibitem[Ottl et~al\mbox{.}(2020)]%
        {Ottl20-GSE}
\bibfield{author}{\bibinfo{person}{Sandra Ottl}, \bibinfo{person}{Shahin
  Amiriparian}, \bibinfo{person}{Maurice Gerczuk}, \bibinfo{person}{Vincent
  Karas}, {and} \bibinfo{person}{Bj\"orn Schuller}.}
  \bibinfo{year}{2020}\natexlab{}.
\newblock \showarticletitle{{Group-level Speech Emotion Recognition Utilising
  Deep Spectrum Features}}. In \bibinfo{booktitle}{\emph{{Proceedings of the
  8th ICMI 2020 EmotiW -- Emotion Recognition In The Wild Challenge (EmotiW
  2020), 22nd ACM International Conference on Multimodal Interaction (ICMI
  2020)}}}. ACM, \bibinfo{publisher}{ACM}, \bibinfo{address}{Utrecht, The
  Netherlands}, \bibinfo{pages}{821--826}.
\newblock


\bibitem[Park et~al\mbox{.}(2022)]%
        {park2022towards}
\bibfield{author}{\bibinfo{person}{Ho-min Park}, \bibinfo{person}{Ilho Yun},
  \bibinfo{person}{Ajit Kumar}, \bibinfo{person}{Ankit~Kumar Singh},
  \bibinfo{person}{Bong~Jun Choi}, \bibinfo{person}{Dhananjay Singh}, {and}
  \bibinfo{person}{Wesley De~Neve}.} \bibinfo{year}{2022}\natexlab{}.
\newblock \showarticletitle{Towards Multimodal Prediction of Time-Continuous
  Emotion Using Pose Feature Engineering and a Transformer Encoder}. In
  \bibinfo{booktitle}{\emph{Proceedings of the 3rd International on Multimodal
  Sentiment Analysis Workshop and Challenge}} (Lisboa, Portugal)
  \emph{(\bibinfo{series}{MuSe' 22})}. \bibinfo{publisher}{Association for
  Computing Machinery}, \bibinfo{address}{New York, NY, USA},
  \bibinfo{pages}{47–54}.
\newblock
\showISBNx{9781450394840}
\urldef\tempurl%
\url{https://doi.org/10.1145/3551876.3554807}
\showDOI{\tempurl}


\bibitem[Pepino et~al\mbox{.}(2021)]%
        {pepino21_interspeech}
\bibfield{author}{\bibinfo{person}{Leonardo Pepino}, \bibinfo{person}{Pablo
  Riera}, {and} \bibinfo{person}{Luciana Ferrer}.}
  \bibinfo{year}{2021}\natexlab{}.
\newblock \showarticletitle{{Emotion Recognition from Speech Using wav2vec 2.0
  Embeddings}}. In \bibinfo{booktitle}{\emph{Proc. Interspeech 2021}}. ISCA,
  \bibinfo{publisher}{ISCA}, \bibinfo{address}{Brno, Czechia},
  \bibinfo{pages}{3400--3404}.
\newblock
\urldef\tempurl%
\url{https://doi.org/10.21437/Interspeech.2021-703}
\showDOI{\tempurl}


\bibitem[Pires et~al\mbox{.}(2019)]%
        {pires-etal-2019-multilingual}
\bibfield{author}{\bibinfo{person}{Telmo Pires}, \bibinfo{person}{Eva
  Schlinger}, {and} \bibinfo{person}{Dan Garrette}.}
  \bibinfo{year}{2019}\natexlab{}.
\newblock \showarticletitle{How Multilingual is Multilingual {BERT}?}. In
  \bibinfo{booktitle}{\emph{Proceedings of the 57th Annual Meeting of the
  Association for Computational Linguistics}}. \bibinfo{publisher}{Association
  for Computational Linguistics}, \bibinfo{address}{Florence, Italy},
  \bibinfo{pages}{4996--5001}.
\newblock
\urldef\tempurl%
\url{https://doi.org/10.18653/v1/P19-1493}
\showDOI{\tempurl}


\bibitem[Priego-Valverde et~al\mbox{.}(2018)]%
        {priego2018smiling}
\bibfield{author}{\bibinfo{person}{B{\'e}atrice Priego-Valverde},
  \bibinfo{person}{Brigitte Bigi}, \bibinfo{person}{Salvatore Attardo},
  \bibinfo{person}{Lucy Pickering}, {and} \bibinfo{person}{Elisa Gironzetti}.}
  \bibinfo{year}{2018}\natexlab{}.
\newblock \showarticletitle{Is smiling during humor so obvious? a
  cross-cultural comparison of smiling behavior in humorous sequences in
  american english and french interactions}.
\newblock \bibinfo{journal}{\emph{Intercultural Pragmatics}}
  \bibinfo{volume}{15}, \bibinfo{number}{4} (\bibinfo{year}{2018}),
  \bibinfo{pages}{563--591}.
\newblock


\bibitem[Sadvilkar and Neumann(2020)]%
        {sadvilkar-neumann-2020-pysbd}
\bibfield{author}{\bibinfo{person}{Nipun Sadvilkar} {and} \bibinfo{person}{Mark
  Neumann}.} \bibinfo{year}{2020}\natexlab{}.
\newblock \showarticletitle{{P}y{SBD}: Pragmatic Sentence Boundary
  Disambiguation}. In \bibinfo{booktitle}{\emph{Proceedings of Second Workshop
  for NLP Open Source Software (NLP-OSS)}}. \bibinfo{publisher}{Association for
  Computational Linguistics}, \bibinfo{address}{Online},
  \bibinfo{pages}{110--114}.
\newblock
\urldef\tempurl%
\url{https://www.aclweb.org/anthology/2020.nlposs-1.15}
\showURL{%
\tempurl}


\bibitem[Schroff et~al\mbox{.}(2015)]%
        {schroff2015facenet}
\bibfield{author}{\bibinfo{person}{Florian Schroff}, \bibinfo{person}{Dmitry
  Kalenichenko}, {and} \bibinfo{person}{James Philbin}.}
  \bibinfo{year}{2015}\natexlab{}.
\newblock \showarticletitle{FaceNet: A unified embedding for face recognition
  and clustering}. In \bibinfo{booktitle}{\emph{2015 IEEE Conference on
  Computer Vision and Pattern Recognition (CVPR)}}. \bibinfo{publisher}{IEEE}.
\newblock
\urldef\tempurl%
\url{https://doi.org/10.1109/cvpr.2015.7298682}
\showDOI{\tempurl}


\bibitem[Serengil and Ozpinar(2020)]%
        {serengil2020lightface}
\bibfield{author}{\bibinfo{person}{Sefik~Ilkin Serengil} {and}
  \bibinfo{person}{Alper Ozpinar}.} \bibinfo{year}{2020}\natexlab{}.
\newblock \showarticletitle{LightFace: A Hybrid Deep Face Recognition
  Framework}. In \bibinfo{booktitle}{\emph{2020 Innovations in Intelligent
  Systems and Applications Conference (ASYU)}}. IEEE, \bibinfo{pages}{23--27}.
\newblock
\urldef\tempurl%
\url{https://doi.org/10.1109/ASYU50717.2020.9259802}
\showDOI{\tempurl}


\bibitem[Stappen et~al\mbox{.}(2021a)]%
        {stappen2021muse}
\bibfield{author}{\bibinfo{person}{Lukas Stappen}, \bibinfo{person}{Alice
  Baird}, \bibinfo{person}{Lukas Christ}, \bibinfo{person}{Lea Schumann},
  \bibinfo{person}{Benjamin Sertolli}, \bibinfo{person}{Eva-Maria Messner},
  \bibinfo{person}{Erik Cambria}, \bibinfo{person}{Guoying Zhao}, {and}
  \bibinfo{person}{Bj{\"o}rn~W Schuller}.} \bibinfo{year}{2021}\natexlab{a}.
\newblock \showarticletitle{The MuSe 2021 multimodal sentiment analysis
  challenge: sentiment, emotion, physiological-emotion, and stress}.
\newblock In \bibinfo{booktitle}{\emph{Proceedings of the 2nd on Multimodal
  Sentiment Analysis Challenge}}. \bibinfo{publisher}{Association for Computing
  Machinery}, \bibinfo{address}{New York, NY, USA}, \bibinfo{pages}{5--14}.
\newblock


\bibitem[Stappen et~al\mbox{.}(2020)]%
        {stappen2020muse1}
\bibfield{author}{\bibinfo{person}{Lukas Stappen}, \bibinfo{person}{Alice
  Baird}, \bibinfo{person}{Georgios Rizos}, \bibinfo{person}{Panagiotis
  Tzirakis}, \bibinfo{person}{Xinchen Du}, \bibinfo{person}{Felix Hafner},
  \bibinfo{person}{Lea Schumann}, \bibinfo{person}{Adria Mallol-Ragolta},
  \bibinfo{person}{Bjoern~W. Schuller}, \bibinfo{person}{Iulia Lefter},
  \bibinfo{person}{Erik Cambria}, {and} \bibinfo{person}{Ioannis
  Kompatsiaris}.} \bibinfo{year}{2020}\natexlab{}.
\newblock \showarticletitle{MuSe 2020 Challenge and Workshop: Multimodal
  Sentiment Analysis, Emotion-Target Engagement and Trustworthiness Detection
  in Real-Life Media}. In \bibinfo{booktitle}{\emph{Proceedings of the 1st
  International on Multimodal Sentiment Analysis in Real-Life Media Challenge
  and Workshop}}. ACM, \bibinfo{publisher}{Association for Computing
  Machinery}, \bibinfo{address}{New York, NY, USA}, \bibinfo{pages}{35–44}.
\newblock


\bibitem[Stappen et~al\mbox{.}(2021b)]%
        {stappen2021summary}
\bibfield{author}{\bibinfo{person}{Lukas Stappen}, \bibinfo{person}{Eva-Maria
  Meßner}, \bibinfo{person}{Erik Cambria}, \bibinfo{person}{Guoying Zhao},
  {and} \bibinfo{person}{Björn~W. Schuller}.}
  \bibinfo{year}{2021}\natexlab{b}.
\newblock \showarticletitle{MuSe 2021 Challenge: Multimodal Emotion,
  Sentiment,Physiological-Emotion, and Stress Detection}. In
  \bibinfo{booktitle}{\emph{29th ACM International Conference on Multimedia
  (ACMMM)}} (Virtual Event, China). ACM, \bibinfo{publisher}{Association for
  Computing Machinery}, \bibinfo{address}{New York, NY, USA}.
\newblock


\bibitem[Stappen et~al\mbox{.}(2021c)]%
        {stappen2021toolbox}
\bibfield{author}{\bibinfo{person}{Lukas Stappen}, \bibinfo{person}{Lea
  Schumann}, \bibinfo{person}{Benjamin Sertolli}, \bibinfo{person}{Alice
  Baird}, \bibinfo{person}{Benjamin Weigel}, \bibinfo{person}{Erik Cambria},
  {and} \bibinfo{person}{Bj{\"o}rn~W Schuller}.}
  \bibinfo{year}{2021}\natexlab{c}.
\newblock \showarticletitle{MuSe-Toolbox: The Multimodal Sentiment Analysis
  Continuous Annotation Fusion and Discrete Class Transformation Toolbox}. In
  \bibinfo{booktitle}{\emph{Proceedings of the 2nd Multimodal Sentiment
  Analysis Challenge, co-located with the 29th ACM International Conference on
  Multimedia (ACMMM)}}. ACM, \bibinfo{publisher}{Association for Computing
  Machinery}, \bibinfo{address}{Changu, China}, \bibinfo{pages}{75--82}.
\newblock


\bibitem[Vaiani et~al\mbox{.}(2022)]%
        {vaiani2022viper}
\bibfield{author}{\bibinfo{person}{Lorenzo Vaiani}, \bibinfo{person}{Moreno
  La~Quatra}, \bibinfo{person}{Luca Cagliero}, {and} \bibinfo{person}{Paolo
  Garza}.} \bibinfo{year}{2022}\natexlab{}.
\newblock \showarticletitle{ViPER: Video-Based Perceiver for Emotion
  Recognition}. In \bibinfo{booktitle}{\emph{Proceedings of the 3rd
  International on Multimodal Sentiment Analysis Workshop and Challenge}}
  (Lisboa, Portugal) \emph{(\bibinfo{series}{MuSe' 22})}.
  \bibinfo{publisher}{Association for Computing Machinery},
  \bibinfo{address}{New York, NY, USA}, \bibinfo{pages}{67–73}.
\newblock
\showISBNx{9781450394840}
\urldef\tempurl%
\url{https://doi.org/10.1145/3551876.3554806}
\showDOI{\tempurl}


\bibitem[Vlasenko et~al\mbox{.}(2021)]%
        {vlasenko2021fusion}
\bibfield{author}{\bibinfo{person}{Bogdan Vlasenko},
  \bibinfo{person}{RaviShankar Prasad}, {and} \bibinfo{person}{Mathew
  Magimai.-Doss}.} \bibinfo{year}{2021}\natexlab{}.
\newblock \showarticletitle{Fusion of Acoustic and Linguistic Information using
  Supervised Autoencoder for Improved Emotion Recognition}.
\newblock In \bibinfo{booktitle}{\emph{Proceedings of the 2nd on Multimodal
  Sentiment Analysis Challenge}}. \bibinfo{publisher}{Association for Computing
  Machinery}, \bibinfo{address}{New York, NY, USA}, \bibinfo{pages}{51--59}.
\newblock


\bibitem[Wagner et~al\mbox{.}(2023)]%
        {wagner2022dawn}
\bibfield{author}{\bibinfo{person}{J. Wagner}, \bibinfo{person}{A.
  Triantafyllopoulos}, \bibinfo{person}{H. Wierstorf}, \bibinfo{person}{M.
  Schmitt}, \bibinfo{person}{F. Burkhardt}, \bibinfo{person}{F. Eyben}, {and}
  \bibinfo{person}{B.~W. Schuller}.} \bibinfo{year}{2023}\natexlab{}.
\newblock \showarticletitle{Dawn of the Transformer Era in Speech Emotion
  Recognition: Closing the Valence Gap}.
\newblock \bibinfo{journal}{\emph{IEEE Transactions on Pattern Analysis \&
  Machine Intelligence}} \bibinfo{number}{01} (\bibinfo{year}{2023}),
  \bibinfo{pages}{1--13}.
\newblock
\showISSN{1939-3539}
\urldef\tempurl%
\url{https://doi.org/10.1109/TPAMI.2023.3263585}
\showDOI{\tempurl}


\bibitem[Weisberg et~al\mbox{.}(2011)]%
        {weisberg2011gender}
\bibfield{author}{\bibinfo{person}{Yanna~J Weisberg}, \bibinfo{person}{Colin~G
  DeYoung}, {and} \bibinfo{person}{Jacob~B Hirsh}.}
  \bibinfo{year}{2011}\natexlab{}.
\newblock \showarticletitle{Gender differences in personality across the ten
  aspects of the Big Five}.
\newblock \bibinfo{journal}{\emph{Frontiers in psychology}}
  \bibinfo{volume}{2} (\bibinfo{year}{2011}), \bibinfo{pages}{178}.
\newblock


\bibitem[Wu et~al\mbox{.}(2021)]%
        {wu2021mumor}
\bibfield{author}{\bibinfo{person}{Jiaming Wu}, \bibinfo{person}{Hongfei Lin},
  \bibinfo{person}{Liang Yang}, {and} \bibinfo{person}{Bo Xu}.}
  \bibinfo{year}{2021}\natexlab{}.
\newblock \showarticletitle{MUMOR: A Multimodal Dataset for Humor Detection in
  Conversations}. In \bibinfo{booktitle}{\emph{CCF International Conference on
  Natural Language Processing and Chinese Computing}}. Springer,
  \bibinfo{publisher}{Springer}, \bibinfo{address}{Qingdao, China},
  \bibinfo{pages}{619--627}.
\newblock


\bibitem[Xu et~al\mbox{.}(2022)]%
        {xu2022hybrid}
\bibfield{author}{\bibinfo{person}{Haojie Xu}, \bibinfo{person}{Weifeng Liu},
  \bibinfo{person}{Jiangwei Liu}, \bibinfo{person}{Mingzheng Li},
  \bibinfo{person}{Yu Feng}, \bibinfo{person}{Yasi Peng},
  \bibinfo{person}{Yunwei Shi}, \bibinfo{person}{Xiao Sun}, {and}
  \bibinfo{person}{Meng Wang}.} \bibinfo{year}{2022}\natexlab{}.
\newblock \showarticletitle{Hybrid Multimodal Fusion for Humor Detection}. In
  \bibinfo{booktitle}{\emph{Proceedings of the 3rd International on Multimodal
  Sentiment Analysis Workshop and Challenge}} (Lisboa, Portugal)
  \emph{(\bibinfo{series}{MuSe' 22})}. \bibinfo{publisher}{Association for
  Computing Machinery}, \bibinfo{address}{New York, NY, USA},
  \bibinfo{pages}{15–21}.
\newblock
\urldef\tempurl%
\url{https://doi.org/10.1145/3551876.3554802}
\showDOI{\tempurl}


\bibitem[Yadav et~al\mbox{.}(2022)]%
        {yadav2022comparing}
\bibfield{author}{\bibinfo{person}{Sarthak Yadav}, \bibinfo{person}{Tilak
  Purohit}, \bibinfo{person}{Zohreh Mostaani}, \bibinfo{person}{Bogdan
  Vlasenko}, {and} \bibinfo{person}{Mathew Magimai.-Doss}.}
  \bibinfo{year}{2022}\natexlab{}.
\newblock \showarticletitle{Comparing Biosignal and Acoustic Feature
  Representation for Continuous Emotion Recognition}. In
  \bibinfo{booktitle}{\emph{Proceedings of the 3rd International on Multimodal
  Sentiment Analysis Workshop and Challenge}} (Lisboa, Portugal)
  \emph{(\bibinfo{series}{MuSe' 22})}. \bibinfo{publisher}{Association for
  Computing Machinery}, \bibinfo{address}{New York, NY, USA},
  \bibinfo{pages}{37–45}.
\newblock
\showISBNx{9781450394840}
\urldef\tempurl%
\url{https://doi.org/10.1145/3551876.3554812}
\showDOI{\tempurl}


\bibitem[Zhang et~al\mbox{.}(2016)]%
        {zhang2016mtcnn}
\bibfield{author}{\bibinfo{person}{Kaipeng Zhang}, \bibinfo{person}{Zhanpeng
  Zhang}, \bibinfo{person}{Zhifeng Li}, {and} \bibinfo{person}{Yu Qiao}.}
  \bibinfo{year}{2016}\natexlab{}.
\newblock \showarticletitle{Joint Face Detection and Alignment Using Multitask
  Cascaded Convolutional Networks}.
\newblock \bibinfo{journal}{\emph{IEEE Signal Processing Letters}}
  \bibinfo{volume}{23} (\bibinfo{date}{04} \bibinfo{year}{2016}).
\newblock


\bibitem[Zhi et~al\mbox{.}(2020)]%
        {zhi2020comprehensive}
\bibfield{author}{\bibinfo{person}{Ruicong Zhi}, \bibinfo{person}{Mengyi Liu},
  {and} \bibinfo{person}{Dezheng Zhang}.} \bibinfo{year}{2020}\natexlab{}.
\newblock \showarticletitle{A comprehensive survey on automatic facial action
  unit analysis}.
\newblock \bibinfo{journal}{\emph{The Visual Computer}}  \bibinfo{volume}{36}
  (\bibinfo{year}{2020}), \bibinfo{pages}{1067--1093}.
\newblock


\bibitem[Zhou and De~la Torre(2015)]%
        {zhou2015generalized}
\bibfield{author}{\bibinfo{person}{Feng Zhou} {and} \bibinfo{person}{Fernando
  De~la Torre}.} \bibinfo{year}{2015}\natexlab{}.
\newblock \showarticletitle{Generalized canonical time warping}.
\newblock \bibinfo{journal}{\emph{IEEE Transactions on Pattern Analysis and
  Machine Intelligence}} \bibinfo{volume}{38}, \bibinfo{number}{2}
  (\bibinfo{year}{2015}), \bibinfo{pages}{279--294}.
\newblock


\end{thebibliography}

\end{document}